\documentclass[letterpaper, 10 pt, conference]{ieeeconf}  % Comment this line out if you need a4paper

\IEEEoverridecommandlockouts                              % This command is only needed if 
\usepackage{xcolor}
\usepackage{pifont}
\usepackage{scalerel}
\usepackage{array}
\usepackage{amsmath}
\usepackage{algpseudocode}   
\usepackage{algorithm}
\usepackage{booktabs}
\usepackage{multirow}
\usepackage{tcolorbox}
\usepackage{listings}
\usepackage{cite}
\usepackage{rotating}
\usepackage{graphicx}
\usepackage{cuted}
\usepackage{caption}
\usepackage{hyperref}
\usepackage{tcolorbox}
\usepackage{subcaption}
\usepackage{afterpage}
\usepackage{inconsolata}
\usepackage[T1]{fontenc}

\definecolor{light-gray}{rgb}{0.8, 0.8, 0.8}
\definecolor{comment-green}{rgb}{0.435, 0.576, 0.106}
\definecolor{prompt-blue}{HTML}{2596be}
\definecolor{code-function}{HTML}{379fbe}
\definecolor{code-function}{HTML}{693da8}  % brian (maybe remove)
\definecolor{code-syntax}{HTML}{0060b1}
\definecolor{code-constant}{HTML}{d86001}
\definecolor{code-comment}{RGB}{106,115,125} % gray-ish
\definecolor{prompt-gray}{HTML}{a7a7a7}
\definecolor{highlight}{HTML}{f8f9cb}
\definecolor{highlight}{HTML}{e3eeff}  % brian (maybe remove)
\definecolor{code-perception}{HTML}{2ecc71}
\definecolor{code-control}{HTML}{ff9900}
\definecolor{code-undefined}{HTML}{ff0000}
\renewcommand\fbox{\fcolorbox{light-gray}{white}}

\newcommand{\hlcode}[1]{\colorbox{highlight}{\makebox[0.99\linewidth][l]{#1}}}

\NewDocumentCommand{\code}{v}{%
\texttt{\small{\textcolor{code-syntax}{#1}}}%
}

\newcommand{\cmark}{{\color{orange}\ding{51}}} % Green check
\newcommand{\xmark}{{\color{black}\ding{55}}}   % Red X

\newcommand{\rot}[1]{\rotatebox{7}{#1}}
\newcommand{\TAMP}[0]{\textsc{ViPR}}
\newcommand{\ours}[0]{\textsc{AnyTask}}
\newcommand{\RL}[0]{\textsc{ViPR-RL}}
\newcommand{\Eureka}[0]{\textsc{ViPR-Eureka}}

\title{\LARGE \bf
\ours{}: an Automated Task and Data Generation Framework for Advancing Sim-to-Real Policy Learning
}

 \author{Ran Gong$^{1*}$, Xiaohan Zhang$^{1*}$, Jinghuan Shang$^{1*}$, Maria Vittoria Minniti$^{1*}$, \\Jigarkumar Patel$^{1}$, Valerio Pepe$^{1}$, Riedana Yan$^{1}$, Ahmet Gundogdu$^{1}$, Ivan Kapelyukh$^{1}$, Ali Abbas$^{1}$, \\ Xiaoqiang Yan$^{1}$, Harsh Patel$^{1}$, Laura Herlant$^{1}$, Karl Schmeckpeper$^{1}$
 \thanks{* Equal Contribution} \\
 \thanks{$^{1}$Robotics and AI Institute, Boston, MA, USA.
        {\tt\small \{rgong, xzhang, jshang, mminniti, jpatel, vpepe, ryan, agundogdu, IKapelyukh, aabbas, xyan, hapatel, lherlant, kschmeckpeper\}@rai-inst.com}}
 }

\begin{document}
% \let\thefootnote\relax\footnotetext{* Equal contribution. }

% \begin{strip}
%   \centering
%   % \vspace{-1.3cm}
%   \includegraphics[width=0.85\textwidth]{figures_and_tables/teaser_draft_copy.pdf}
%   \captionof{figure}{\ours{} is a framework that automates task design and generates data for robot learning. The resulting data enables training visuomotor policies that can be deployed directly onto a physical robot without requiring any real-world data. }
%   \label{fig:title_figure}
% \end{strip}

% \maketitle

\twocolumn[{%
  \renewcommand\twocolumn[1][]{#1}%
  \maketitle
  \begin{center}
    \centering
    \vspace{-1.2cm} % Fine-tune the gap between authors and figure here
    \includegraphics[width=0.85\textwidth]{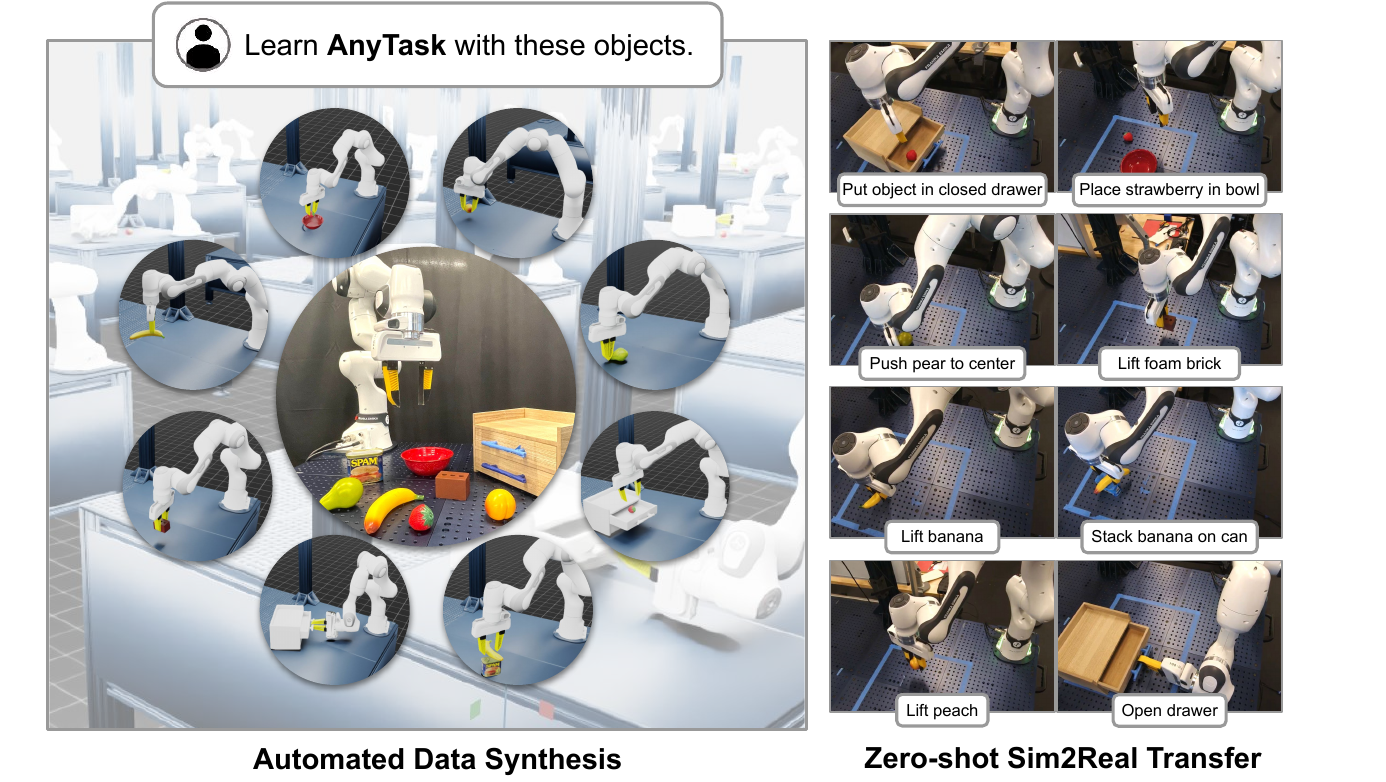}
    \captionof{figure}{\ours{} is a framework that automates task design and generates data for robot learning. The resulting data enables training visuomotor policies that can be deployed directly onto a physical robot without requiring any real-world data.}
    \label{fig:title_figure}
  \end{center}%
}]

{
  % Switch to symbols
  \renewcommand{\thefootnote}{\fnsymbol{footnote}}
  % Force the footnote index to 1 (which maps to *)
  \footnotetext[1]{Equal Contribution}
}
\footnotetext[1]{The authors are with the Robotics and AI Institute, Boston, MA, USA.
{\tt\small \{rgong, xzhang, jshang, mminniti, jpatel, vpepe, ryan, agundogdu, IKapelyukh, aabbas, xyan, hapatel, lherlant, kschmeckpeper\}@rai-inst.com}}

\vspace{-0.8\baselineskip}
\thispagestyle{empty}
\pagestyle{empty}

%%%%%%%%%%%%%%%%%%%%%%%%%%%%%%%%%%%%%%%%%%%%%%%%%%%%%%%%%%%%%%%%%%%%%%%%%%%%%%%%
\begin{abstract}
Generalist robot learning remains constrained by data: large-scale, diverse, and high‐quality interaction data are expensive to collect in the real world. 
% Simulation becomes a promising way for scaling up data collection, but still, simulation task design, task-aware scene generation, expert demonstration synthesis, and sim-to-real transfer all demand substantial human effort. 
While simulation has become a promising way for scaling up data collection, the related tasks, including simulation task design, task-aware scene generation, expert demonstration synthesis, and sim-to-real transfer, still demand substantial human effort.
We present \ours{}, an automated framework that pairs massively parallel GPU simulation with foundation models to design diverse manipulation tasks and synthesize robot data. 
We introduce three \ours{} agents for generating expert demonstrations aiming to solve as many tasks as possible: 1) \TAMP{}, a novel task and motion planning agent with VLM-in-the-loop Parallel Refinement; 2) \Eureka{}, a reinforcement learning agent with generated dense rewards and LLM-guided contact sampling; 3) \RL{}, a hybrid planning and learning approach that jointly produces high-quality demonstrations with only sparse rewards.
We train behavior cloning policies on generated data, validate them in simulation, and deploy them directly on real robot hardware. 
The policies generalize to novel object poses, achieving 44\% average success across a suite of real-world pick-and-place, drawer opening, contact-rich pushing, and long-horizon manipulation tasks. Our project website is at \href{https://anytask.rai-inst.com}{https://anytask.rai-inst.com}.
\end{abstract}

%%%%%%%%%%%%%%%%%%%%%%%%%%%%%%%%%%%%%%%%%%%%%%%%%%%%%%%%%%%%%%%%%%%%%%%%%%%%%%%%
\section{Introduction}
% The success of deep learning is fundamentally linked to the access to large-scale, high-quality data, ~\cite{deng2009imagenet,toshniwal2024openmath2, schuhmann2022laion}, demonstrated in various domains such as language modeling~\cite{achiam2023gpt,touvron2023llama, liu2024deepseek, bai2023qwen},  visual understanding~\cite{he2016deep,dosovitskiy2021an,tschannen2025siglip,shang2024theia, heinrich2025radiov2,radford2021learning,alayrac2022flamingo}, and generation~\cite{rombach2021highresolution,li2024dreamscene, chen2023scenedreamer},  and multimodal applications~\cite{liu2023visual,shridhar2022cliport,li2024llara}. 

The success of deep learning fundamentally depends on access to large-scale, high-quality data \cite{deng2009imagenet,toshniwal2024openmath2, schuhmann2022laion}, as demonstrated in various domains such as language modeling \cite{achiam2023gpt,touvron2023llama, liu2024deepseek, bai2023qwen}, visual understanding \cite{he2016deep,dosovitskiy2021an,tschannen2025siglip,shang2024theia, heinrich2025radiov2,radford2021learning,alayrac2022flamingo}, generation \cite{rombach2021highresolution,li2024dreamscene, chen2023scenedreamer}, and multimodal applications \cite{liu2023visual,shridhar2022cliport,li2024llara}. However, collecting robot data is extremely time-consuming and costly~\cite{black2024pi0,galaxeag0} as it necessitates direct physical interaction with the real world. 
% Robot simulation~\cite{mujoco,coppeliasim,robosuite2020}, which can be scaled straightforwardly by computing~\cite{isaacsim,isaaclab,xiang2020sapien}, 
Robot simulation, which can be scaled straightforwardly with compute~\cite{isaacsim,isaaclab,xiang2020sapien}, 
presents an appealing alternative for collecting large-scale datasets with minimal real-world effort~\cite{rlbench,gong2023arnold,liu2023libero,yu2020meta,mu2025robotwin,chen2025robotwin}. %robocasa,robomimic,
While prior work has made significant progress in designing simulation systems for a wide range of tasks, tremendous human effort is often a huge barrier in building these systems~\cite{taomaniskill3, li2024behavior}.
% ~\rutav{`inevitable' suggests that this is never possible, but you are making it possible in this paper. maybe some other word choice?}. 
This effort includes proposing tasks, selecting task-relevant object assets, designing metrics, ensuring feasibility, and generating a large quantity of high-quality demonstration data. 
These non-trivial components frequently limit the diversity of the generated data.

% \begin{figure}
%   \centering
%   \vspace{-0.5cm}
%   \includegraphics[width=0.99\linewidth]{figures_and_tables/teaser_draft_copy.pdf}
%   \captionof{figure}{\ours{} is a framework that automatically designs tasks and generates data for robot learning. The resulting data enables training sim-to-real policies that can be deployed directly onto a physical robot without requiring any real-world data. }
%   \label{fig:title_figure}
%   \vspace{-0.5cm}
% \end{figure}

\begin{figure*}
\vspace{5mm}
    \centering
    \includegraphics[width=0.96\linewidth]{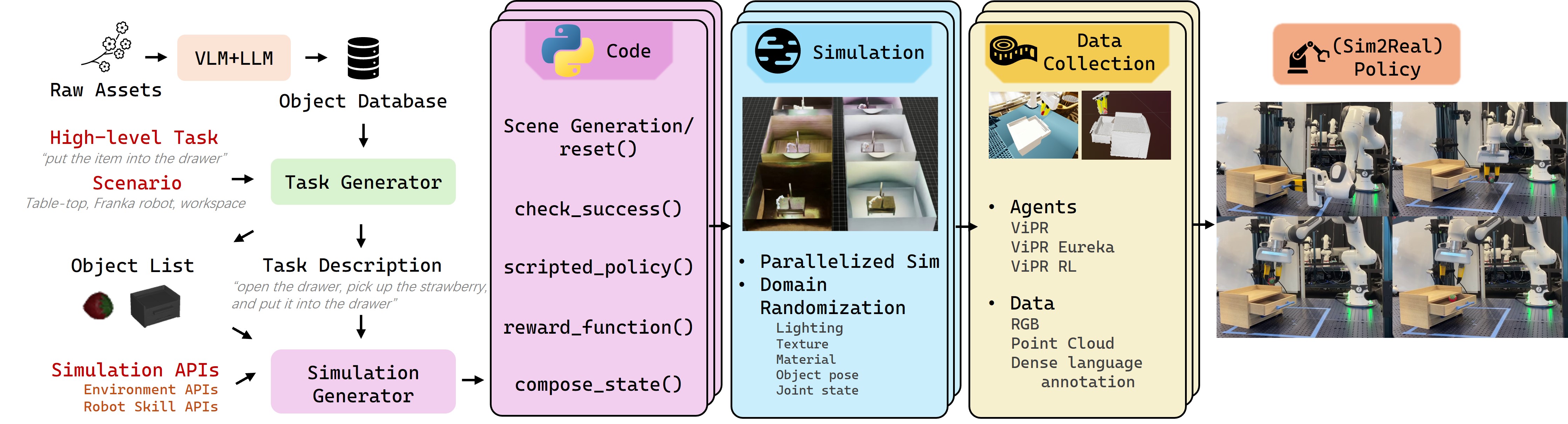}
    \caption{Overview of \ours{}. We first generate simulated manipulation tasks from an object database and a high-level task (i.e., task type). Then the pipeline automatically proposes task descriptions, generates the simulation code, and efficiently collects data using different agents, including \TAMP{}, \RL{}, and \Eureka{} in massively parallel simulation environments. We apply online domain randomization in the simulation to ensure the diversity of the scenes and the visual observations. Finally, we train the policy using simulated data and zero-shot transfer to the real world.}
    \label{fig:system}
    \vspace{-6mm}
\end{figure*}

Trained on vast internet data, foundation models demonstrate remarkable abilities in robotic downstream applications, such as scene understanding, task planning, motion synthesis, and low-level control~\cite{kim2024openvla, liu2023llm+, gu2024conceptgraphs, cui2024anyskill}. 
These capabilities can also be leveraged to automate many key steps in creating robotic simulation environments, such as task design, writing simulation code, and iterative refinement.
% ~\rutav{its not intuitive why? maybe some citations to support this argument}. 
% However, prior work leveraging foundation models for robot simulations has generally focused on discrete sub-problems, such as task generation or demonstration collection. Crucially, these approaches still require significant human intervention, leaving the overall process far from fully automated.
However, prior work leveraging foundation models for robot simulations either requires significant human efforts on task design and demonstration collection~\cite{huagensim2, mu2025robotwin, chen2025robotwin}, or struggles with sim-to-real transfer~\cite{robogen, wang2023gensim}, even though the ultimate goal of large-scale data collection is to deploy the trained system in the real world.
% Furthermore, these efforts are largely confined to the simulation domain, with little discussion of the sim-to-real transfer, even though the ultimate goal of large-scale data collection is to deploy the trained system in the real world.

To address the aforementioned challenges, we introduce \ours{} (\autoref{fig:title_figure}), a scalable framework designed to bridge the gap between current simulators and a fully automated data generation system. 
The primary goal of \ours{} is to leverage massively parallel GPU-based simulation engines and foundation models to generate high-quality, diverse scenes, tasks, and expert demonstrations at scale. 
To achieve this, our framework (as illustrated in~\autoref{fig:system}) integrates an intelligent object database, a task generator, and a simulation generator, all orchestrated by LLMs, to produce diverse, large-scale manipulation datasets for robust sim-to-real transfer (\autoref{sec:anytask}).
To synthesize robotic trajectories for a diverse set of tasks, we introduce three \ours{} agents built upon task and motion planning (TAMP) and reinforcement learning (RL): \TAMP{}, \Eureka{}, and \RL{}. This data is used to train visuomotor policies that are directly deployable in the real world (\autoref{sec:methodology}). 
Furthermore, we design a task management workflow and a demonstration replay mechanism to accelerate the data collection process (\autoref{sec:infra}).
The entire pipeline, from task generation to policy training, operates almost autonomously, requiring only a high-level textual objective.

In summary, we make the following contributions: \begin{itemize}
    \item  We present \ours{}, a novel, automated framework that leverages massively parallel, GPU-based simulation to acquire robotic data from high-level goals, significantly reducing the need for manual intervention.
    \item Based on the highly parallel nature of our framework, we introduce \TAMP{}, \Eureka{}, and \RL{} agents that can automatically generate expert demonstrations on \ours{} at scale.
    \item We validate the utility of our generated data by training and evaluating visuomotor policies on a suite of manipulation tasks in simulation.
    \item We demonstrate \textit{zero-shot} policy transfer to a physical robot, and identify key factors, such as domain randomization and policy architecture, that are critical for bridging the sim-to-real gap.
\end{itemize}

\section{Related Works}
\begin{table*}[t!]
    \vspace{3mm}
    \centering
    \resizebox{0.95\linewidth}{!}{
\begin{tabular}{lccccccccccc}
\toprule
            &  \rot{Auto Task Generation} & \rot{Auto Trajectory Generation } & \rot{Auto Object} & \rot{Dense Annotation} & \rot{Task Metric Generation} & \rot{Scene Generation} & \rot{Domain Randomization} & \rot{Massively Parallel GPU Simulation} & \rot{Long Horizon} & \rot{ZeroShot Perceptual Sim-to-real Transfer} & \rot{Rendering}   \\
\midrule
    RoboGen \cite{wang2023robogen}  & \cmark  & RL, TAMP, H & \cmark  & \xmark & \cmark  & \cmark & \xmark & \xmark & \xmark & \xmark & R  \\
    Gen2Sim \cite{katara2024gen2sim} & \cmark & RL & \cmark  & \xmark & \cmark & \cmark & \xmark & \xmark & \xmark & \xmark & R \\
    GenSim \cite{wang2023gensim} &  \cmark & TAMP & \xmark  &  \xmark & \cmark & \cmark & \xmark & \xmark & \cmark &   \xmark & R \\ 
    GenSim2 \cite{huagensim2} & \cmark  & \xmark & \xmark  & \xmark & \cmark  & \cmark & \xmark & \xmark & \xmark &  \cmark & R\\
    ScalingUp \cite{ha2023scaling} & \cmark  & TAMP & \xmark  & \xmark & \cmark  & \cmark & \xmark & \xmark & \cmark &  \cmark & R \\ 
    RoboTwin \cite{mu2025robotwin, chen2025robotwin} &  \xmark  & TAMP & \cmark & \xmark  & \cmark  & \cmark & \xmark & \xmark & \xmark &  \cmark & R \\
    Meta-world \cite{yu2020meta} & \xmark  & RL & \xmark  & \xmark & \xmark  & \cmark & \xmark & \xmark & \xmark &  \xmark & R \\
    Robocasa \cite{robocasa2024} & \cmark  & \xmark & \cmark  & \xmark & \cmark  & \cmark & \xmark & \xmark & \cmark &  \xmark & R \\ 
    ARNOLD \cite{gong2023arnold} & \xmark  & TAMP & \xmark  & \xmark & \xmark  & \cmark & \xmark & \xmark & \xmark &  \cmark & RT \\
    Maniskill \cite{taomaniskill3, gu2023maniskill2} & \xmark  & RL, TAMP & \xmark  & \xmark & \xmark  & \xmark & \xmark & \cmark & \xmark &  \cmark  & RT\\
    LIBERO \cite{liu2023libero} & \xmark  & \xmark & \xmark  & \xmark & \xmark  & \cmark & \xmark & \xmark & \xmark &  \xmark & R \\
   % AutoBio \cite{lan2025autobio} & \xmark  & \cmark & \xmark  & \xmark & \xmark  & \cmark & \xmark & \xmark & \xmark &  \xmark \\
   % SkillGen \cite{pmlr-v270-garrett25a} & \xmark & \xmark & \xmark & \\
    %GRUTopia \cite{wang2024grutopia} & \cmark  & \cmark & \xmark  & \xmark & \xmark  & \cmark & \xmark & \xmark & \xmark  & \xmark & RT \\ 
    Behavior1K \cite{li2024behavior} & \xmark & \xmark & \xmark & \xmark & \xmark & \xmark & \xmark & \cmark & \cmark & \xmark & RT \\
\midrule
    AnyTask (ours) & \cmark & RL, TAMP, H & \cmark & \cmark & \cmark & \cmark & \cmark & \cmark & \cmark & \cmark & RT\\
\bottomrule

\end{tabular}
}
\caption{Comparison with other simulation systems. \textbf{Auto Task Generation}: Automatic task generation from a single text prompt. \textbf{Auto Trajectory Generation}: Automatic trajectory generation with no human effort. \textbf{RL: } Using reinforcement learning to generate demonstrations. \textbf{TAMP: } Using task and motion planning approach to generate demonstrations. \textbf{H}: Combing TAMP and RL in a single trajectory. \textbf{Auto Object}: Automatic object creation/indexing with no human effort. \textbf{Dense Annotation}: Dense annotation for each robot manipulation. \textbf{Task Metric Generation}: Task Metric Generation. \textbf{Scene Generation}: Automatic Scene Generation. \textbf{Domain Randomization}: Automatic Online Physical Visual Domain Randomization. \textbf{Massively Parallel GPU Simulation}: Massively parallel GPU-based simulation with massively parallel cameras to support scalable task, scene, and trajectory generation as well as domain randomization. \textbf{Long Horizon}: Long horizon task and demo generation. \textbf{ZeroShot Perceptual Sim-to-real Transfer}: Zero-shot Perceptual sim-to-real for a closed-loop policy. \textbf{RT}: Real-time ray tracing. \textbf{R}: Rasterization or non-real-time ray tracing.  } 

\label{tab:sim_comparsion}
\vspace{-5mm}
\end{table*}

\subsection{Large-Scale Robotics Dataset in Simulation}

 Recent progress in simulations enabled large-scale robot data collection \cite{deng2025graspvla, taomaniskill3, gu2023maniskill2, mees2022calvin}; however, most tasks are still manually curated. This process requires substantial human effort to design, implement, and validate meaningful tasks. More recently, several studies have explored using large language models (LLMs) to automatically propose tasks~\cite{wang2023robogen, huagensim2, wang2023gensim, katara2024gen2sim, ha2023scaling}. However, these efforts typically do not focus on scaling to larger datasets, addressing sim-to-real transfer, or developing diverse and systematic data collection strategies.
 
In contrast, we introduce a holistic, end-to-end pipeline designed to automate the entire data generation lifecycle while directly addressing the sim-to-real challenge, as shown in \autoref{tab:sim_comparsion}. Our system integrates asset selection, scene configuration, task generation, task success criterion generation, policy data collection, policy distillation, and real-world deployment, all with significantly reduced human efforts.

\subsection{Sim-to-Real Transfer}
Recent years have witnessed impressive advancements in sim-to-real transfer \cite{tang2023industreal, handa2023dextreme, akkaya2019solving, yu2024natural}. However, these methods often rely on meticulously human-designed reward functions or complex training pipelines. A promising direction involves leveraging Large Language Models (LLMs); for instance, recent work has demonstrated the feasibility of using LLM-generated tasks for sim-to-real \cite{huagensim2}. Our work demonstrates that by using LLMs to generate not only the tasks but also the data collection policies, we can achieve competitive sim-to-real performance. While concurrent work \cite{chen2025robotwin} also achieves zero-shot sim-to-real transfer, their approach relies on policies pre-trained with real-world robot data. In contrast, we establish that effective real-world policy transfer is achievable using a framework built upon purely synthetic data for a diverse range of tasks.

%Data generation through teleoperation in sim
%\cite{mandlekar2023mimicgen}

%ProcTHOR~\cite{deitke2022} - generative environments

%BBSEA
%\cite{yang2024bbsea}

%Robogen
%\cite{wang2023robogen}

%GenSim
%\cite{wang2023gensim}
%- only does top-down manipulation tasks

%GenSim2
%\cite{huagensim2}
%- only does 24 tasks
%no controlled reset
%predefined criteria

%RoboTwin \cite{mu2025robotwin} - 17 tasks; generation happens at 1) 2D RGB image %to 3D model and texture, and 2) code for data collection

%RoboTwin2 \cite{chen2025robotwin}

%Some methods for evaluating different generative simulation methods. \cite{chen2024evaluation}
% input: task type, skill apis, a set of env apis
% output: <tsm, description, observation function, dense reward function>, initial TAMP plan, scene <assets, domain randomization, dense annotation>

\section{\ours{}}
\label{sec:anytask}
\ours{} aims to generate text-based task descriptions and corresponding runnable simulation code for agents to collect synthetic data. The system overview is available in \autoref{fig:system}. In the sections below, we introduce Object Database, Task Generator, Simulation Generator, and other key infrastructure components.
% The task and scene generation pipeline is automated, with options to specify tasks, objects, and extra instructions. 
% \ours{} is built on top of a parallel physics simulation and a promptable object database.
% Given a prompt of high-level task family (e.g., ``pick-and-place'', ``stacking objects'', or ``opening the drawer''), and a set of callable environment and robot skill Python APIs, \ours{} can synthesize a suite of simulated tasks as well as agents to accomplish the tasks through our task generator and simulation generator. Below, we introduce the generators and the agents, respectively.

\paragraph{Object Database}
We build an object database storing objects' information so that retrieving objects through natural language is possible.
% It also gives enough object metadata for code generation. 
The object database is built based on the available assets before task generation. 
% Given assets in Universal Scene Description (USD) format, we traverse the tree-like structure and extract the information as fine-grained as at the part level. 
The database encodes objects and object parts for their names, colors, textures, materials, bounding boxes (extent), joint information (for articulated objects), and overall descriptions. This process combines textual and visual information by rendering each object and part from multiple viewpoints, and asking a VLM (GPT-4o) to give the annotations related to visual properties. 
% We particularly instruct the VLM to give an accurate number of the major parts (like the number of drawers in one cabinet), which is useful during retrieval and composing policy. 
% The object database also stores the properties of all the joints. 
We use Sentence-T5-Large~\cite{raffel2020exploring,reimers-2019-sentence-bert} to compute sentence-level embeddings, and then use \texttt{faiss}~\cite{douze2024faiss, johnson2019billion} to build an index for nearest neighbor search. Therefore, no human efforts are involved with object database creation and query, beyond finding assets. Examples of labeled object metadata can be found in appendix~\ref{sec:appendix_object_database}.

\paragraph{High-level task and scenario information} Our goal is to generate a diverse set of realistic and physically plausible robotic tasks. To achieve this, we prompt an LLM with high-level information, including a task family (e.g., ``pick-and-place''), robot specifications, and workspace constraints, all provided in natural language by a human.

\paragraph{Task Generator}
Task generator uses the high-level task and scenario information to propose tasks and objects with the help of the object database. We support two variants. In \textbf{object-based task generation}, objects are first sampled from a database, and the LLM then proposes a detailed task involving them. This is flexible for general tasks like ``pick-and-place''. In \textbf{task-based object proposal}, the LLM first suggests objects suitable for a given task (e.g., a drawer for ``open a drawer''). The system then retrieves a matching asset from the object database, and the LLM generates the final, detailed task description. This stage outputs a natural language task description and structured object metadata, including the object details in the object database.

\paragraph{Simulation Generator} Our simulation generator takes the task description and objects as input and produces code that can execute based on our simulation framework. To execute the generated code and leverage massively GPU simulations, we use IsaacLab~\cite{isaaclab} simulator. We choose to generate code to define a task since code has less ambiguity than natural language and has higher flexibility than configuration files. We provide environment and robot skill API definitions as part of the prompt. 
The APIs are designed by humans to ensure correctness. API lists are available in~\autoref{tab:appendix_env_apis} and \autoref{tab:appendix_skill_apis}.
% For skill APIs, we design them to be as generic as possible, so we leverage LLMs to chain them to accomplish complicated tasks.

%In simulation code generation, the LLM is required to design the following four functions that are necessary for executing the task: \texttt{reset()}, \texttt{check\_success()}, \texttt{compose\_state()}, and \texttt{scripted\_policy()}. \texttt{reset()} is responsible for resetting and randomizing the environment after each epoch, mainly about placing objects, which gives many more variants of the scene compared to generating a static configuration file. We resolve object collisions by pushing the later placed object away from the colliding object if there is one. \texttt{check\_success()} checks if the current environment is in a successful state. \texttt{compose\_state()} provides the environment state for training a state-based RL policy, where we encourage it to include as much of the prevailing environment information for a quicker and easier RL process. \texttt{scripted\_policy()} defines the task motion planning policy as another data collection option in parallel to the RL policy. The scripted policy uses the provided skill APIs for the motion with the help of other APIs in a way like code-as-policy~\cite{}. 

In detail, the LLM is required to generate the code for five key functions: \texttt{reset()} for randomizing the scene (e.g., object poses), \texttt{check\_success()} to define the task's goal condition, \texttt{compose\_state()} to provide a state representation for an RL policy, \texttt{reward\_function()} for an initial version of reward function in RL, and \texttt{scripted\_policy()} to define an expert policy for data collection. To ensure these functions are consistent, we generate \texttt{check\_success()} first and use it to instruct the LLM when generating the other four functions.

\paragraph{Dense Annotation}
Language annotations are limited in existing robot datasets~\cite{black2024pi0,galaxeag0}, %robocasa
where only one or a few sentences of task description are often paired with one demonstration. We introduce our automated dense annotation system to bridge this gap.
We transform the privileged information in the simulation into dense, natural language annotations to summarize the environmental states before and after executing an action. Each annotation is tagged to a certain timestep or a period of time of the trajectory. 
To generate the annotations, we provide an API \texttt{log\_step()} so that the LLM can call it any time during policy execution, and decide what information to include using other environment APIs. 
% In the prompt provided to LLM, we primarily encourage LLM to summarize the state before and after executing a scripted policy. 
In this way, we can automatically generate data with rich text annotations, providing strong support for our policy refinement (introduced later).
An example dense annotation is shown below, where the variables will be replaced by the actual value in the simulation.
% In addition, we provide dense annotation programs, structured as shown below, providing text-based context for our agents.

{\scriptsize
\begin{verbatim}
{
  'step': 0,
  'content': {
  'step_description':{'action':'Move end-effector to drawer handle',..},
  'object_states': {'drawer_handle': {'position': [x,y,z], ...}, ...},
  'robot_state': {'eef_pos': [x,y,z], ...} 
  }, ...
}
\end{verbatim}
}
A more integrated example can be found in \autoref{fig:vipr_policy}.

\section{\ours{} Agents}
\label{sec:methodology}
% We present three different agents that solve generated tasks from \ours{} and collect expert demonstrations: \TAMP{}, a novel Task and motion planning TAMP agent with \textbf{V}LM-\textbf{i}n-the-loop \textbf{P}arallel \textbf{R}efinement; \Eureka{}, an improved version of Eureka~\cite{ma2024eureka} with VLM-fintuned sparse reward and Mesh-based Contact Sampling; and \RL{}, a hybrid TAMP+RL approach.

This section describes the agents we developed and evaluated for \ours{}.  
Our guiding principle is to \textit{explore how a robot agent can solve as many generated tasks as possible with no human effort.}

In this work, we study task and motion planning (TAMP) and reinforcement learning (RL), two commonly used teacher policies in manipulation tasks. TAMP is known for handling long-horizon tasks, while RL excels at dexterous, contact-rich manipulation. 
However, traditional TAMP methods require pre-defined domain and action knowledge, usually specified in PDDL format by domain experts. 
RL, on the other hand, is typically limited to specific domains where practitioners can carefully construct reward functions that provide accurate learning signals for the desired behavior.

To this end, we introduce three \ours{} agents for generating expert demonstrations:  
\begin{itemize}
    \item \TAMP{}, a novel TAMP agent with \textbf{V}LM-\textbf{i}n-the-loop \textbf{P}arallel \textbf{R}efinement,  
    \item \Eureka{}, an improved version of Eureka~\cite{ma2024eureka} with VLM-finetuned sparse rewards and Mesh-based Contact Sampling, and  
    \item \RL{}, a hybrid TAMP+RL approach.  
\end{itemize}

\subsection{\TAMP{}}
A \TAMP{} agent uses an LLM to produce a task–motion plan \(p\) as a Python program that calls our parameterized skill APIs (in our case, \texttt{move\_to}, \texttt{open\_gripper}, \texttt{close\_gripper}, \texttt{grasp}), following prior approaches on code generation for robot control~\cite{liang2022code}. 
% Running such programs open-loop often fails due to limited spatial understanding in foundation models~\cite{openeqa}, leading to inaccurate end-effector poses.
Naively running the generated programs in an open-loop manner often leads to failure, mainly because LLMs (and foundation models in general) lack robust spatial understanding of the environment~\cite{openeqa}. A common failure mode is commanding inaccurate 3D positions or orientations for the end-effector.

% \noindent\textbf{Policy improvement via \emph{parallel} VLM feedback.}
To mitigate this limitation, we propose to use VLMs for iteratively refining the task-motion plan.
Each refinement iteration takes as input: the current plan \(p\), images collected during rollout, dense annotations from \ours{}, and the available skill and environment APIs. The iteration outputs an updated plan \(p'\).
We execute \(K\) parallel rollouts of \(p\) in simulation to (i) record videos and dense trajectory annotations and (ii) expose diverse failure modes in a single pass. 
A VLM evaluates every such rollout and outputs natural-language feedback plus a binary success/failure with confidence. 
We aggregate these per-episode judgments into a scalar which is the success rate across the \(K\) rollouts combined with average confidence and compare VLM judgments against \texttt{check\_success} that only inspects initial and final states to monitor agreement. A generated example can be found in \autoref{fig:vipr_policy}.

\subsection{\Eureka{}}
To generate demonstrations with reinforcement learning, we use an imporved version of Eureka \cite{ma2024eureka} to iteratively refine and sample reward functions proposed by LLM.

\noindent \textbf{Mesh-based Contact Sampling: }
 A core component of our approach is a novel contact sampling algorithm. The sampler generates high-quality grasp candidates by first sampling a triangle on the object mesh and applying barycentric interpolation \cite{osada2002trimeshsampling} to determine a contact position.

% Formally, let the object mesh consist of N triangles. 
% Let the mesh consist of \(N\) triangles
% \[
%   \tau_i = \bigl(v_{i1},\,v_{i2},\,v_{i3}\bigr), \quad i=1,\dots,N,
% \]

% with area \[ A_i \;=\;\frac12\,\bigl\|\,(v_{i2}-v_{i1})\times(v_{i3}-v_{i1})\bigr\|. \]
% Define a categorical distribution over triangles:
% \[
%   P(I = i) \;=\; \frac{A_i}{\sum_{j=1}^N A_j},
%   \qquad I \sim \mathrm{Cat}\bigl(P(1),\dots,P(N)\bigr).
% \]

% % 2. Barycentric interpolation within the chosen triangle
% Once \(\tau_I\) is selected, sample two random numbers
% \[
%   r_1, r_2 \;\overset{\mathrm{i.i.d.}}{\sim}\; \mathcal{U}(0,1), 
%   \quad u = \sqrt{r_1}.
% \]
% Compute barycentric coordinates:
% \[
%   \lambda_1 = 1 - u,\quad
%   \lambda_2 = u\,(1 - r_2),\quad
%   \lambda_3 = u\,r_2,
%   \quad
%   \sum_{k=1}^{3} \lambda_k = 1.
% \]
% The contact point is
% \[
%   \mathbf{p}
%   = \sum_{k=1}^3 \lambda_k\,v_{I k}
%   = \lambda_1\,v_{I1} + \lambda_2\,v_{I2} + \lambda_3\,v_{I3}.
% \]

The object of interest is first generated by an LLM. The right gripper finger is then positioned along the perturbed surface normal at the sampled location with Gaussian noise. To speed up sampling efficiency, the gripper orientation is randomly sampled around a predefined gripper z-axis produced by a vision language model (VLM) or human users.

\begin{figure*}[ht]
\centering
\includegraphics[width=\linewidth]{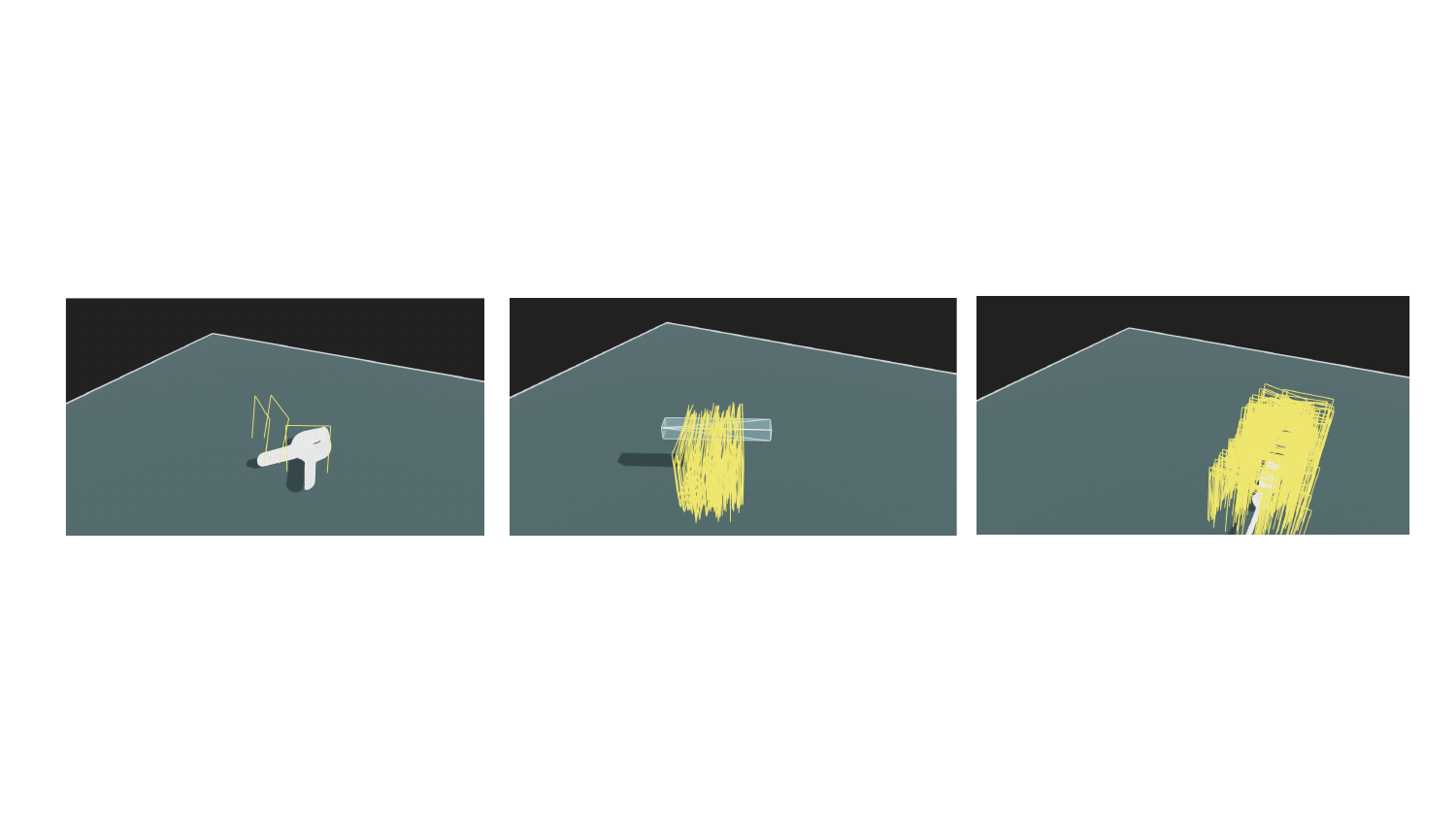}

\caption{Visualizations of contact sampling results for the clamp (left), drawer handle collision mesh (middle), and screwdriver (right). } \label{fig:contactSampling}
\end{figure*}

To ensure feasibility, we employ a rejection sampling mechanism that discards grasp candidates with collisions or invalid orientations based on collision checking. We sample and check around 1024 candidates per environment in parallel using batched collision checking and batched inverse kinematics (IK) with cuRobo \cite{sundaralingam2023curobo}. The resulting valid position is then used as the initial state for RL training. After the training success rate improves, we gradually decay the number of environments using contact sampling. Some example contact sampling results are shown in \autoref{fig:contactSampling} and an example \RL{} policy can be found in \autoref{fig:rl_policy}.

\subsection{\RL{}}
We want to combine the strengths of both worlds, RL and motion planning, since RL is good with contact-rich tasks, and motion planning is good at free space movement. Several modifications are needed: 1) The code generation now includes trained RL skills with APIs. 2) For each sub-task, we use motion planning to move the gripper to the object parts of interest, which are sampled by the grasp sampler described above, and then we invoke trained RL skills.  To train an object-based RL skill, we run the PPO 1500 epochs with 1024 environments for each object; it typically requires 20 minutes on an L4 GPU for a single object.  The reward function is a simple success checker produced by LLM. A sample code snippet is shown below.

\vspace{7mm}
\noindent\fbox{\parbox{0.97\linewidth}
{\scriptsize{
\texttt{{%
{\color{prompt-gray}%
% {\color{code-syntax}import} torch\\
{\color{code-syntax}def} {\color{code-function}ViPR\_policy\_rl}(env):}\\
%\query{{\color{code-comment}\# Step 1: Use an RL skill to grasp the baseball}}\\
\hlcode{{\color{code-comment}\hspace*{4mm}\# Object IDs}}\\
\hlcode{\hspace*{4mm}baseball\_id = {\color{code-constant}1}}\\
% \hlcode{\hspace*{4mm}pitcher\_base\_id = {\color{code-constant}2}}\\
\hlcode{{\color{code-comment}\hspace*{4mm}\# Get the grasp pose from the RL skill API}}\\
\hlcode{\hspace*{4mm}grasp\_position, grasp\_orientation = {\color{code-function}get\_grasp\_position}(}\\
\hlcode{\hspace*{8mm}env, baseball\_id, part\_name=""}\\
\hlcode{\hspace*{4mm})}\\
\hlcode{{\color{code-comment}\hspace*{4mm}\# Move to the grasp pose}}\\
\hlcode{\hspace*{4mm}hover\_offset = torch.{\color{code-function}tensor}([[{\color{code-constant}0.0}, {\color{code-constant}0.0}, {\color{code-constant}0.1}]])}\\
\hlcode{\hspace*{4mm}above\_grasp = grasp\_position + hover\_offset}\\
\hlcode{\hspace*{4mm}{\color{code-function}move\_to}(env=env, target\_position=above\_grasp,}\\
\hlcode{\hspace*{12mm}target\_orientation=grasp\_orientation, gripper\_open={\color{code-constant}True})}\\
\hlcode{\hspace*{4mm}{\color{code-function}move\_to}(env=env,target\_position=grasp\_position, }\\
\hlcode{\hspace*{12mm}target\_orientation=grasp\_orientation, gripper\_open={\color{code-constant}True})}\\
% \hlcode{\hspace*{12mm}target\_orientation=grasp\_orientation, gripper\_open={\color{code-constant}True})}\\
\hlcode{{\color{code-comment}\hspace*{4mm}\# Execute the RL picking skill}}\\
 \hlcode{\hspace*{4mm}pick\_success = {\color{code-function}pick\_rl}(env, external\_id=baseball\_id)}\\
% \hlcode{\hspace*{4mm}pitcher\_position = {\color{code-function}get\_object\_position}(env, pitcher\_base\_id)}\\
% \hlcode{{\color{code-comment}\hspace*{4mm}\# Define the target placement position}}\\
% \hlcode{\hspace*{4mm}place\_offset = torch.{\color{code-function}tensor}([[{\color{code-constant}0.0}, {\color{code-constant}0.0}, {\color{code-constant}0.05}]])}\\
% \hlcode{\hspace*{4mm}target\_position = pitcher\_position + place\_offset}\\
% \hlcode{\hspace*{4mm}above\_target = target\_position + hover\_offset}\\
% \hlcode{{\color{code-comment}\hspace*{4mm}\# Move to the placement pose and release}}\\
% \hlcode{\hspace*{4mm}{\color{code-function}move\_to}(env=env, target\_position=above\_target,}\\
% \hlcode{\hspace*{12mm}target\_orientation={\color{code-constant}None}, gripper\_open={\color{code-constant}False})}\\
% \hlcode{\hspace*{4mm}{\color{code-function}move\_to}(env=env, target\_position=target\_position,}\\
% \hlcode{\hspace*{12mm}target\_orientation={\color{code-constant}None}, gripper\_open={\color{code-constant}False})}\\
% ...\\
% \hlcode{\hspace*{4mm}{\color{code-function}open\_gripper}(env)}\\
\hlcode{\hspace*{4mm}{\color{code-function}move\_to}(env=env, target\_position=above\_grasp,}\\
\hlcode{\hspace*{12mm}target\_orientation={\color{code-constant}None}, gripper\_open={\color{code-constant}False})}\\
}}}}}

A more integrated example can be found in \autoref{fig:combined_rl_logic} in the appendices.

% \begin{itemize}
%     \item  State Replay: We directly set the simulation to the stored states of a successful trajectory. %By replaying only successful attempts, we bypass the time spent on failed rollouts, which we hypothesize provides a significant speed-up over standard methods.
%     \item Action Replay: We re-execute the original action sequence from a successful trajectory within the physics engine. %This process tests the robustness of the actions, allowing us to filter out sequences that may fail under minor variations in the physics engine's non-determinism.
% \end{itemize}

% \paragraph{Data Saver}
% \todo{Xiaohan}

\section{Infrastructure Design}
\label{sec:infra}
\subsection{Multi-GPU Data Collection}
The data collection pipeline is orchestrated using Metaflow~\cite{metaflow} to manage the sequential execution of each stage and the data artifacts produced within a simulation environment.
% The process is structured into distinct stages to maximize flexibility and computational efficiency; this modular design is inherently scalable, allowing the framework to manage data generation for a large number of distinct tasks. 
There are three stages. It begins with an optional policy refinement stage, allowing for the enhancement of an agent's performance prior to data collection. In the second stage, the primary data gathering is conducted using a state-based policy, which efficiently captures a diverse range of interaction trajectories without rendering. In the final stage, these collected trajectories are replayed to render and capture high-fidelity vision data. This decoupling of collection logic from the rendering process significantly reduces computational overhead and allows for independent iteration on visual parameters.  We launch a Metaflow pipeline on each GPU node so data can be collected in parallel. The resulting Metaflow-managed agents provide a fast and adaptable workflow for generating large-scale, vision-based datasets.

\subsection{Demonstration Replay}
Instead of generating and recording demonstrations at the same time, we first execute a state-based policy, (\TAMP{}, \Eureka{} or \RL{} ) to generate numerous rollouts in parallel without recording. We then store the states from only the successful trajectories. These successful trajectories are replayed to train our final policy. We employ two replay methods: 1) State Replay: We directly set the simulation to the stored states of a successful trajectory. 2)  Action Replay: We re-execute the original action sequence from a successful trajectory.

\section{Experiments}
We perform experiments to evaluate our code generation~(\ref{sec:code_runability}) and the diversity~(\ref{sec:language_diversity}) of our tasks, the success rates~(\ref{sec:data_gen_success}) and speed~(\ref{sec:data_gen_speed}) of data generation, and the performance of policies trained on our data in simulation~(\ref{sec:simulation_success}) and the real world~(\ref{sec:sim2real_success}).

\subsection{Are programs synthesized by \ours{} runnable in simulation?}
\label{sec:code_runability}
\begin{table}[ht]
    \centering
    \vspace{3mm}
    \caption{Code runability analysis.}
    \label{tab:tab_taskgen_runability}
    \resizebox{1.0\linewidth}{!}{%
        \begin{tabular}{lcccc}
            \toprule
            \multirow{2}{*}{LLM} & \multirow{2}{*}{Runnable} & \multicolumn{3}{c}{Error Type} \\
            \cmidrule(lr){3-5}
            & & \texttt{compose\_state()} & \texttt{reset()} & \texttt{check\_success()} \\ 
            \midrule
            o1-mini & 64\% & 0\% & 20\% & 16\% \\
            DeepSeek-R1~\cite{guo2025deepseek} & 76\% & 0\% & 16\% & 8\% \\
            o3-mini & 84\% & 4\% & 12\% & 0\% \\
            o3-mini + improved prompt & 96\% & 4\% & 0\% & 0\% \\
            \bottomrule
        \end{tabular}%
    }
    \vspace{-3mm}
\end{table}
% \todo{Jinghuan -- add text}
\ours{} relies on LLM to generate code that runs in the simulation. We tested several LLMs: o1-mini, DeepSeek-R1-671B~\cite{guo2025deepseek}, and o3-mini by using them to generate 20 tasks with the same set of objects. The test only focuses on the basic simulation environment loop, not the policy, so only \texttt{compose\_state()}, \texttt{reset()}, and \texttt{check\_success()} are executed. We report the code runability -- the ratio of the code that can run in simulation, and in which functions errors may frequently occur. \autoref{tab:tab_taskgen_runability} reports the code runability. We find that o3-mini gives the highest code runability. We also find that the errors often come from the \texttt{reset()} function, since that function requires strong logic to handle object placement and spatial transforms. We further summarize the errors in these tests and compose an improved prompt targeted to those errors. With the improved prompt, \ours{} can achieve a code runability of 96\% using o3-mini.

% \subsection{Are success checkers from \ours{} aligned with human judgment?}
% \label{sec:success_checkers}
% \input{figures_and_tables/human_evaluated_success_checker}
% We sampled five tasks from each task family for human evaluation. The results, as shown in \autoref{tab:human_eval}, demonstrates that our generated success checker, while not perfect, is highly aligned with human judgment.

\subsection{How diverse is \ours{} compared to other data generation systems in the literature?}
\label{sec:language_diversity}
\begin{table}[H]
    \centering
    \caption{BLEU Score of Generated Task Descriptions}
    \begin{tabular}{lccc}
    \toprule
    Ours     & RoboGen~\cite{wang2023robogen} & RLBench~\cite{rlbench} & Gensim2~\cite{huagensim2} \\
    0.352     & 0.494 & 0.590 & 0.692 \\ 
    \bottomrule
    \end{tabular}
    \label{tab:bleu_score}
\end{table}
Diversity is one of the key aspects of data quality. We use self-BLEU score~\cite{papineni2002bleu} to evaluate the diversity of the generated task descriptions. We compare our system against RoboGen~\cite{wang2023robogen}, RLBench~\cite{rlbench}, and GenSim2~\cite{huagensim2}. Since our system requires human input for high-level tasks, we use the high-level manipulation tasks from RoboGen~\cite{wang2023robogen} to generate our detailed task descriptions. We compute the self-BLEU score of the task descriptions from each method using \texttt{n\_grams=4}. The result is available in \autoref{tab:bleu_score}. Our system has the lowest self-BLEU score, showing that our task descriptions have better diversity than other methods.

\subsection{How robust are \ours{} agents on generating expert demonstrations?}
\label{sec:data_gen_success}
% \begin{figure}[H]
%     \centering
%     \includegraphics[width=1.0\linewidth]{figures_and_tables/data_gen_stats.png}
%     \caption{Data generation success rate across task types \todo{combine with ~\ref{tab:agents_success}}}
%     \label{fig:data_gen_stats}
% \end{figure}
We collect data using \TAMP{} across five task categories: \textbf{lifting}, \textbf{pick-and-place}, \textbf{pushing}, \textbf{stacking}, and \textbf{drawer opening} with varying difficulties, totaling more than 400 tasks. 
%For each category, our generation framework instantiates tasks with progressively increasing numbers of distracting objects. 
%For example, \texttt{lifting\_2obj} corresponds to a lifting task with one distractor, while \texttt{lifting\_3obj} includes two distractors.

As shown in \autoref{tab:agents_success}, each agent excels at different tasks, enabling the ensemble to collectively solve more tasks than any single agent. This highlights the necessity of agent diversity, as certain approaches are ill-suited for specific agents. 
Qualitatively, \Eureka{} is able to learn to grasp a grasping complex object, like a bleach bottle, while the other methods fail because they cannot explore enough to find the single viable angle.  \RL{} can solve a stacking task that requires knocking over one of the objects before stacking the second object on top, while \TAMP{} cannot learn to knock over the object and \Eureka{} struggles with the multi-step nature of the task. Finally, \TAMP{} is most successful at multi-step tasks that do not require unique behaviors.
\begin{table}[h]
\centering
\caption{Percentage of tasks that \ours{} agents can successfully solve (i.e., success rate >10\%).} 
\resizebox{\linewidth}{!}{%

\begin{tabular}{lcccccc}
\toprule
& Lifting & Pushing & Stacking & Pick \& Place & Drawer opening \\
\midrule
\TAMP{} w/o Refinement &  58\% &  30\% &  26\% &   66\%  &  0\%  \\
\TAMP{} & \textbf{81}\% &  54\% & \textbf{44}\% &  \textbf{76}\%  &  0\% &    \\
\RL{} &   35\% &  --\% &   9\% &  33\%  & \textbf{33}\%  \\
\Eureka{} &  69\% &   \textbf{60}\% &  33\% & 73\% &  17\%  \\
All &  90\% &  70\% & 54\% & 87\% & 33\% \\
\bottomrule
\end{tabular}%
}
\label{tab:agents_success}
\end{table}

% For instance, we did not train the \RL{} agent on pushing because defining a generalizable pushing skill is difficult.

% \noindent \textit{Lifting and Pick-and-Place.} 
% These tasks achieve a relatively higher success rate of {>73\%}. 
% Failures primarily arise when sampled objects are not physically graspable, making the task unsolvable.

% \noindent \textit{Pushing.} 
% Pushing tasks are more challenging, with performance dropping to around {50\%}. 
% Certain objects (e.g., spoons or clamps) tend to move laterally when pushed, making controlled pushing less reliable.

% \noindent \textit{Stacking.} 
% Stacking tasks achieve a success rate of {38\%}. 
% These tasks are inherently more difficult as they require long-horizon planning with multiple sequential pick-and-place operations. 
% Failures are more common because sampled objects may not be pickable, and the current framework does not ensure that one object can rest stably on another.

% \noindent \textit{Drawer Opening.} 
% Drawer opening achieves 65.17\% average success rate for data collection. 

%\noindent \textit{Pattern Creation.} 
%Pattern creation tasks achieve a success rate of \textcolor{blue}{32\%}. These tasks require arranging multiple objects into specific spatial configurations (e.g., circles, squares, or lines). Performance is limited by the compounding of picking and placement errors across multiple attempts.

% \todo{Xiaohan: Drawer opening data collection success rate}

\subsubsection{Is refinement in \TAMP{} useful?}
\begin{figure}[H]
    \centering
    \includegraphics[width=1.0\linewidth]{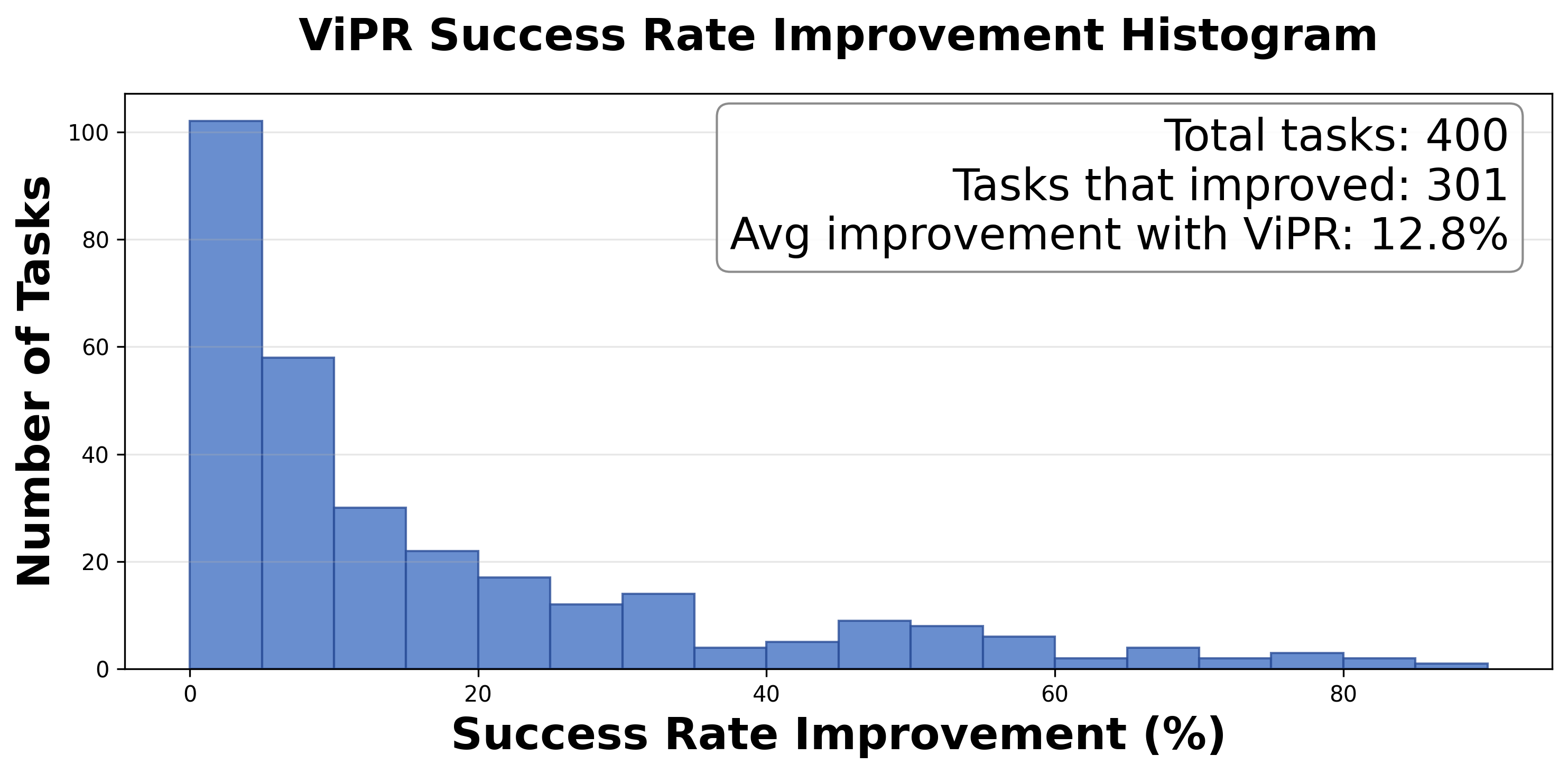}
    \caption{ViPR improvement: Using ViPR leads to an average $12.8\%$ improvement in success rate on 301 tasks}
    \label{fig:task_refinement_improvement}
    \vspace{-5mm}
\end{figure}

The VLM refiner improves success rates in $86.4\%$ of tasks, with an average gain of $13.6\%$ for tasks with non-zero initial success. This consistently produces more robust policies \autoref{fig:task_refinement_improvement}.

\subsubsection{Is contact sampling useful?}
To demonstrate the effectiveness of LLM-guided contact sampling, we perform ablation studies against the vanilla Eureka.

\noindent As shown in \autoref{tab:contact_sampling_rl},  \Eureka{} significantly outperforms the standard Eureka. All experiments are run across \textbf{30} tasks within the task family, with 3 Eureka iterations and 3 tries each. A task was considered successful if any iteration in any try achieved a success rate exceeding 10\% .
\begin{table}[h]
\vspace{2mm}
\centering
\caption{Data collection RL policy training success rate with and without contact sampling}
\setlength{\tabcolsep}{1pt}
\resizebox{\linewidth}{!}{%
\begin{tabular}{lcccccc}
\toprule
& Lifting & Pushing & Stacking & Pick\&Place & DrawerOpening & Avg. \\
\midrule
Eureka & 40 \% & 40 \% & 0 \% &  57 \%  & \textbf{50} \% & 37 \%  \\
\Eureka{} & \textbf{73} \% & \textbf{50} \% & \textbf{57} \% & \textbf{87} \% & 17 \% & \textbf{57} \% \\
\bottomrule
\end{tabular}%
}
\label{tab:contact_sampling_rl}
\end{table}

\subsection{How fast can \ours{} agents generate data?}
\label{sec:data_gen_speed}
The throughput of \ours{} data generation is determined by two key factors: 
the success rate of the AnyTask Agents and the trajectory length (i.e., the number of simulation timesteps) required to complete a task.
To optimize throughput, we decompose the pipeline into two stages: (1) demonstration recording and (2) trajectory replay. 
During the first stage, AnyTask agents attempt tasks without rendering, and only simulator states from successful trajectories are stored. 
In the second stage, the saved simulator states are replayed with rendering enabled to generate the full dataset, 
including RGB images, colored point clouds, robot states, and action sequences required for imitation learning. 
% Our two-stage recording pipeline significantly increases overall data generation throughput, enabling the collection of 500 demonstrations, with 11 seconds each episode per task for \textbf{$\sim$36 minutes} on a single L4 GPU 
Our two-stage recording pipeline is highly efficient. In a single \textbf{$\sim$36 minutes} session on an L4 GPU, we collected 500 demonstrations, recording RGB-D and point cloud data from 4 cameras for each 11-second demo. This total time accounts for all overhead, including instance launching, isaac-sim shader compiling, point cloud computation, data saving and data uploading time.
%(see Fig.\ref{fig:throughput})
% \begin{figure}[H]
%     \centering
%     \includegraphics[width=0.8\linewidth]{figures_and_tables/throughput_analysis.png}
%     \caption{Data generation throughput}
%     \label{fig:throughput}
% \end{figure}
% \input{figures_and_tables/data_generation_speed}

% \subsection{How fast is action replay?}
\begin{figure}
    \vspace{3mm}
    \centering
    \includegraphics[width=0.95\linewidth]{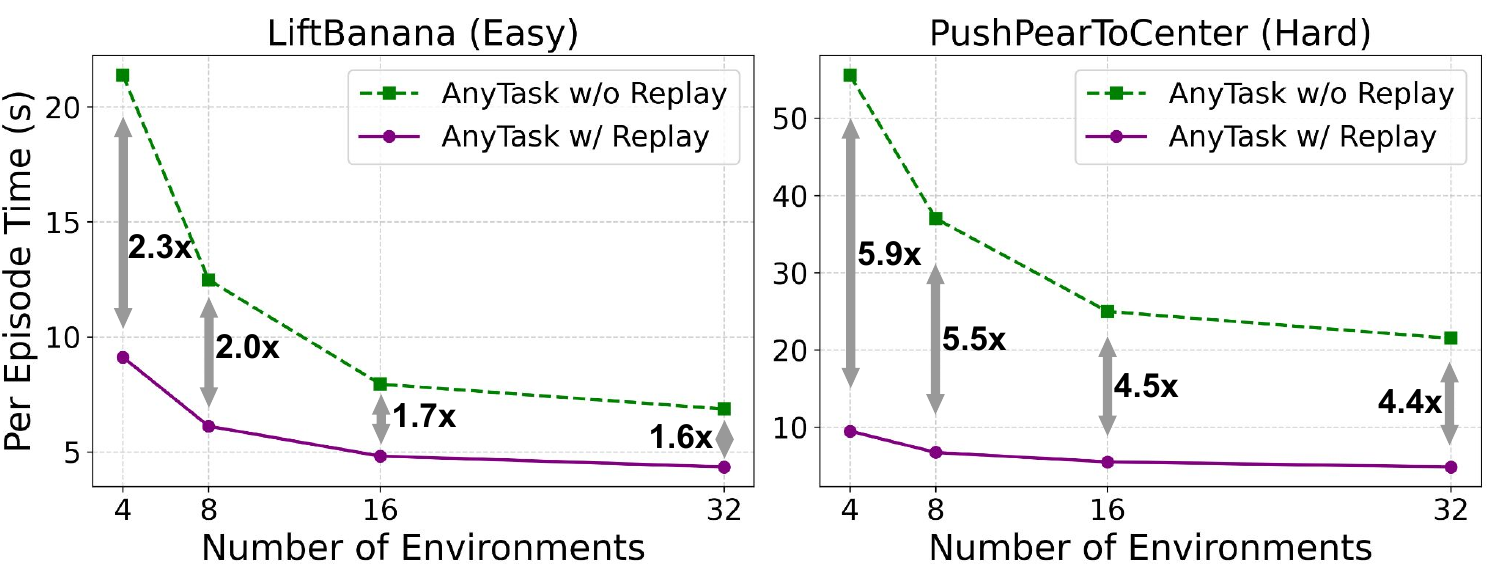}
    \caption{Action replay enables faster data collection, especially on challenging tasks.}
    \label{fig:action_replay}
    \vspace{-8mm}
    
\end{figure}
As shown in \autoref{fig:action_replay}, action replay significantly improves data generation throughput by eliminating wasted rendering time. This benefit is especially pronounced on more difficult tasks, where agents often struggle to generate successful trajectories.  In 4-camera environments, this method resulted in a four-fold speedup on more difficult tasks.

\subsection{Can we train BC policies with data generated by \ours{} agents?}
\label{sec:simulation_success}
\begin{table}[H]
    \vspace{-4mm}
    \centering
    \caption{Policy evaluations in simulation.}
    \resizebox{0.75\columnwidth}{!}{
    \begin{tabular}{cccc}
    \toprule
     Task family &  \TAMP{} &  \RL{} & \Eureka{} \\
     \midrule
     Lifting & 42.0\% & 9.4\% & 4.7\% \\
     Pick and place & 40.7\% &  0.8\% & -\\
     Pushing & 29.3\% &  - & 19.8\% \\
     Stacking*  & 2.0\% & 0\% & 1.7\% \\
     % Object rearrangement & & & \\
     Drawer  & - & 29.1\% & 2.6\% \\

     % & \multicolumn{2}{c}{\textbf{Ours (\TAMP{})}} & \multicolumn{2}{c}{\textbf{Ours (\RL{})}} & \multicolumn{2}{c}{\textbf{Ours (\Eureka{})}}   \\
     % Task family & Data generation & BC & Data generation & BC & Data generation & BC \\
     % \midrule
     % Lifting & 76/100 & 42.0\% & 44/100 & 9.4\% & 69/100 & 4.7\% \\
     % % Lifting - Hard & & 5.7\% & & 0\% & & 0.3\% \\ % 2 object
     % Pick and place & 73/100 & 40.7 \% & 36/100 & 0.8\% & - & -\\
     % Pushing & 47/100 & 29.3\% & - & - & 59/100 & 19.8\% \\
     % Stacking & 37/100 & & & & & \\
     % % Object rearrangement & & & & & & \\
     % Drawer  & - & 2/5 & 29.1\% & 3/5 & \\

    \bottomrule
    \end{tabular}
    }
    \label{tab:sim_success}
    \vspace{-4mm}
% \todo{Xiaohan: Evaluate Drawer opening success rate in sim}
    
\end{table}
We train diffusion policies on each of the generated tasks.  \autoref{tab:sim_success} shows the policy success rates in simulation on a subset of the tasks that all data collection methods successfully generated data for, comparing over 70 policies.  \TAMP{} data performs better on tasks that involve long horizon or multi-step processes, while data collection methods that include RL perform equal or better on tasks that involve continuous contact.  Despite high data collection efficiencies, \Eureka{}  data is more difficult to distill into BC policies. For pick-and-place tasks, it hacks the reward system by pushing objects instead of picking them up. Example task descriptions that policies were successful on are shown in \autoref{tab:task_descriptions}.
\begin{table}[H]
    \centering
    \caption{Example task descriptions}
    \resizebox{\columnwidth}{!}{
    \begin{tabular}{l}
    \toprule

Grasp the strawberry and lift it vertically off the table by about 10 cm, then hold it steady for a few \\

\hspace{12pt}seconds while ensuring the nearby plate remains undisturbed. \\

Pick up the extra large clamp and place it slightly forward (positive x direction) relative to the cup. \\

Push the pear diagonally (forward and left) so that it settles between the racquetball and the fork. \\ 

Stack the baseball on top of the potted meat can, ensuring the baseball is directly aligned above the can. \\

    \bottomrule
    \end{tabular}
    }
    \label{tab:task_descriptions}
\end{table} More details related to policy learning are in the appendix \ref{sec:policy_learning_sup}.

\subsection{Are these policies transferrable to real world?}
\label{sec:sim2real_success}
\begin{figure}
    \vspace{3mm}
    \centering
    \includegraphics[width=.95\linewidth]{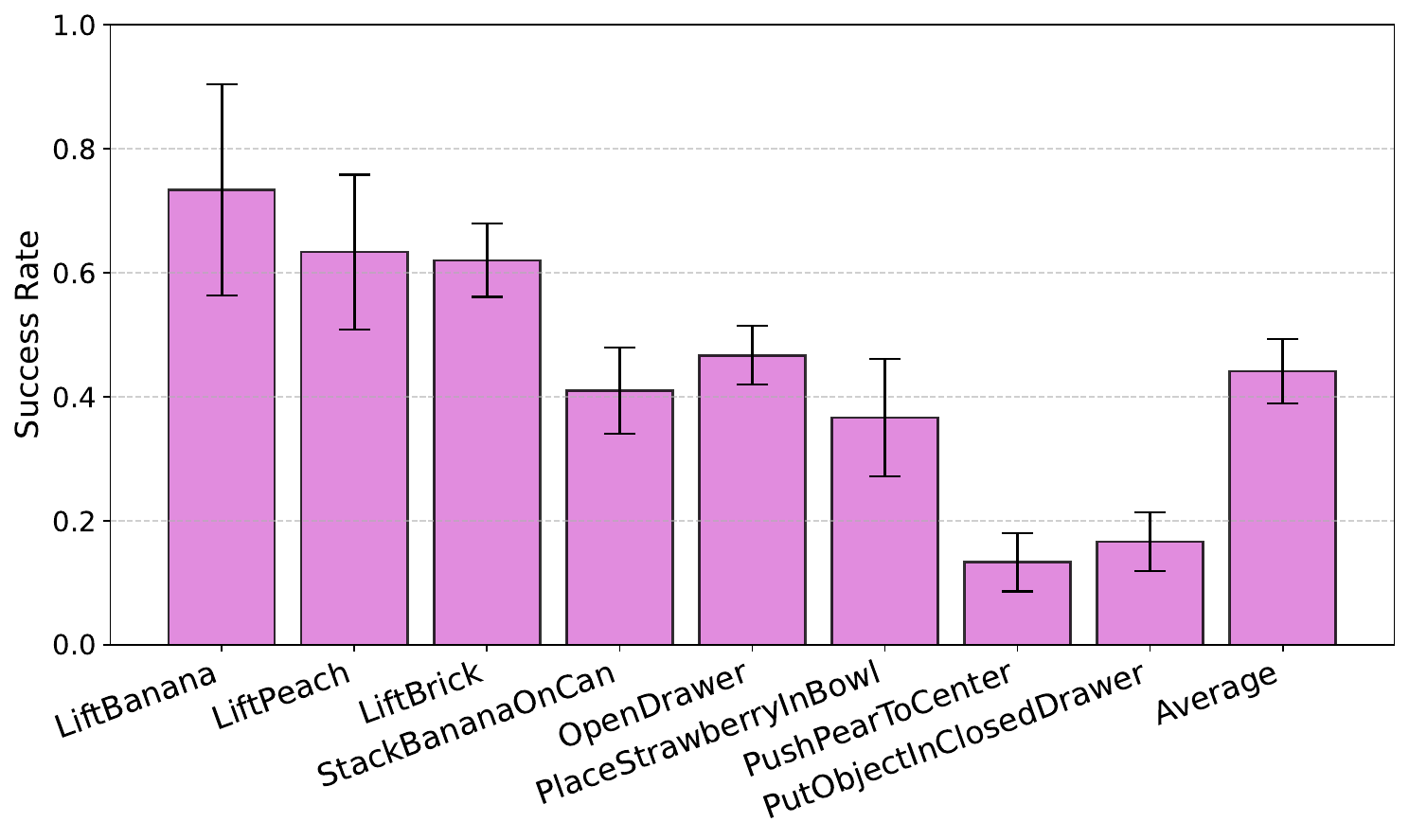}
    \caption{Zero-shot sim-to-real policy evaluations.}
    \label{fig:real}
    \vspace{-3mm}
\end{figure}
We generated eight tasks (see \autoref{fig:real}) in \ours{} and used \TAMP{} to collect 1{,}000 expert demonstrations per task.
These demonstrations vary in length from 10 seconds (e.g., \textit{LiftBanana}) to 30 seconds (e.g., \textit{PutObjectInClosedDrawer}, where the robot opens the drawer, picks up the strawberry, and places it inside).
We distill these demonstrations into a set of single-task, point-cloud--based policies using 3D Diffusion Policy~\cite{DP3}.
Note that for drawer-related tasks, we provide an additional \texttt{open\_drawer} skill API for \TAMP{} to generate high-quality trajectories.
Each single-task policy is trained on 4$\times$NVIDIA H100 GPUs for 500 epochs with a global batch size of 1{,}024.
We use a cosine learning-rate schedule (initial LR $5\times 10^{-5}$) with 100 warm-up iterations and weight decay $1\times 10^{-6}$.

For better sim-to-real performance, we use uncolored point clouds as visual input.
The workspace uses four tabletop RealSense D455 cameras, and we use an image resolution of $320\times240$.
We fuse points from the four cameras and uniformly subsample to $4{,}096$ points.
We crop out table points.
We apply small pose jitter: translations in $[-1,1]$\,cm and rotations in $[-2^\circ,2^\circ]$.
To mimic depth artifacts, we simulate ``ghost'' points: up to $5\%$ of the cloud, $70\%$ biased near object boundaries within a $10\%$ shell; depths include Gaussian noise with $\sigma=3$\,mm.
We use the absolute end-effector pose and a discrete gripper state (0=open, 1=closed) as proprioceptive inputs.

For the action space, the policy predicts a chunk of 64 actions and each action is an absolute end-effector poses and a desired gripper state.
We implement an asynchronous policy runner with temporal ensembling~\cite{act} to execute 32 actions from each predicted chunk.
The policy runs locally on a single A6000 GPU at 30Hz.

We evaluate each of the eight policies for 30 trials with randomly sampled object poses within the workspace.
Error bars are computed by partitioning the 30 trials into three groups of 10 and reporting the mean~$\pm$~s.e.m.\ across groups.
\autoref{fig:real} shows per-task success; the policies show generalization to novel object poses, achieving 44\% average success. More details can be found in appendix \ref{sec:sim-to-real-appendix}.

% \section{CONCLUSIONS, LIMITATIONS and FUTURE WORK}
% We introduced \ours{}, a scalable framework for automated task generation and sim-to-real policy learning that supports multiple data-generation agents, including \TAMP{}, \RL{}, \Eureka{}, leveraging privileged information. We demonstrated the framework's utility through simulation evaluations and, most notably, deployed a policy trained from scratch directly onto a real robot without any real-world fine-tuning.

% While we successfully demonstrated sim-to-real transfer for point-cloud-based policies, an interesting direction for future work is to study the transfer of RGB-based policies. We also plan to scale the framework to include more objects and robot types, as well as extend it to mobile manipulation tasks by synthesizing complex, scene-level object arrangements.

\section{CONCLUSIONS, LIMITATIONS and FUTURE WORK}

In this work, we addressed the critical data bottleneck in robot learning by introducing \ours{}, a framework that automates the entire pipeline from high-level task to sim-to-real policy deployment. We demonstrated how \ours{} leverages foundation models and parallel simulation to automatically generate diverse tasks, scenes, and success criteria. Our novel data generation agents, including the TAMP-based \TAMP{} and RL-based \Eureka{} and \RL{}, efficiently produce high-quality expert demonstrations for a wide range of manipulation challenges. Our approach is validated by training a visuomotor policy purely on this synthetic data and deploying it zero-shot to a physical robot, achieving notable performance across various tasks without any real-world fine-tuning.

Despite these promising results, our framework has several limitations that present exciting avenues for future research. First, while our agents demonstrate broad capabilities, their performance varies on tasks requiring high-precision or complex physical reasoning, such as stacking arbitrary objects. Second, our successful sim-to-real transfer relied on point-cloud observations. Extending this to RGB-based policies would be a valuable direction, as it would lower the barrier for real-world deployment on a wider variety of hardware. We also plan to scale the framework to include a greater diversity of objects and robot morphologies, as well as extend it to more complex, long-horizon mobile manipulation tasks.


%%%%%%%%%%%%%%%%%%%%%%%%%%%%%%%%%%%%%%%%%%%%%%%%%%%%%%%%%%%%%%%%%%%%%%%%%%%%%%%%

%%%%%%%%%%%%%%%%%%%%%%%%%%%%%%%%%%%%%%%%%%%%%%%%%%%%%%%%%%%%%%%%%%%%%%%%%%%%%%%%

%\section*{ACKNOWLEDGMENT}

%%%%%%%%%%%%%%%%%%%%%%%%%%%%%%%%%%%%%%%%%%%%%%%%%%%%%%%%%%%%%%%%%%%%%%%%%%%%%%%%

\bibliography{root}

\begin{thebibliography}{10}

\bibitem{deng2009imagenet}
J.~Deng, W.~Dong, R.~Socher, L.-J. Li, K.~Li, and L.~Fei-Fei, ``Imagenet: A large-scale hierarchical image database,'' in {\em 2009 IEEE conference on computer vision and pattern recognition}, pp.~248--255, Ieee, 2009.

\bibitem{toshniwal2024openmath2}
S.~Toshniwal, W.~Du, I.~Moshkov, B.~Kisacanin, A.~Ayrapetyan, and I.~Gitman, ``Openmathinstruct-2: Accelerating ai for math with massive open-source instruction data,'' {\em arXiv preprint arXiv:2410.01560}, 2024.

\bibitem{schuhmann2022laion}
C.~Schuhmann, R.~Beaumont, R.~Vencu, C.~Gordon, R.~Wightman, M.~Cherti, T.~Coombes, A.~Katta, C.~Mullis, M.~Wortsman, {\em et~al.}, ``Laion-5b: An open large-scale dataset for training next generation image-text models,'' {\em Advances in neural information processing systems}, vol.~35, pp.~25278--25294, 2022.

\bibitem{achiam2023gpt}
J.~Achiam, S.~Adler, S.~Agarwal, L.~Ahmad, I.~Akkaya, F.~L. Aleman, D.~Almeida, J.~Altenschmidt, S.~Altman, S.~Anadkat, {\em et~al.}, ``Gpt-4 technical report,'' {\em arXiv preprint arXiv:2303.08774}, 2023.

\bibitem{touvron2023llama}
H.~Touvron {\em et~al.}, ``Llama: Open and efficient foundation language models,'' {\em arXiv preprint arXiv:2302.13971}, 2023.

\bibitem{liu2024deepseek}
A.~Liu, B.~Feng, B.~Xue, B.~Wang, B.~Wu, C.~Lu, C.~Zhao, C.~Deng, C.~Zhang, C.~Ruan, {\em et~al.}, ``Deepseek-v3 technical report,'' {\em arXiv preprint arXiv:2412.19437}, 2024.

\bibitem{bai2023qwen}
J.~Bai, S.~Bai, Y.~Chu, Z.~Cui, K.~Dang, X.~Deng, Y.~Fan, W.~Ge, Y.~Han, F.~Huang, {\em et~al.}, ``Qwen technical report,'' {\em arXiv preprint arXiv:2309.16609}, 2023.

\bibitem{he2016deep}
K.~He, X.~Zhang, S.~Ren, and J.~Sun, ``Deep residual learning for image recognition,'' in {\em Proceedings of the IEEE Conference on Computer Vision and Pattern Recognition}, pp.~770--778, 2016.

\bibitem{dosovitskiy2021an}
A.~Dosovitskiy, L.~Beyer, A.~Kolesnikov, D.~Weissenborn, X.~Zhai, T.~Unterthiner, M.~Dehghani, M.~Minderer, G.~Heigold, S.~Gelly, J.~Uszkoreit, and N.~Houlsby, ``An image is worth 16x16 words: Transformers for image recognition at scale,'' in {\em International Conference on Learning Representations}, 2021.

\bibitem{tschannen2025siglip}
M.~Tschannen {\em et~al.}, ``Siglip 2: Multilingual vision-language encoders with improved semantic understanding, localization, and dense features,'' {\em arXiv preprint arXiv:2502.14786}, 2025.

\bibitem{shang2024theia}
J.~Shang, K.~Schmeckpeper, B.~B. May, M.~V. Minniti, T.~Kelestemur, D.~Watkins, and L.~Herlant, ``Theia: Distilling diverse vision foundation models for robot learning,'' {\em arXiv preprint arXiv:2407.20179}, 2024.

\bibitem{heinrich2025radiov2}
G.~Heinrich, M.~Ranzinger, H.~Yin, Y.~Lu, J.~Kautz, A.~Tao, B.~Catanzaro, and P.~Molchanov, ``Radiov2.5: Improved baselines for agglomerative vision foundation models,'' in {\em Proceedings of the Computer Vision and Pattern Recognition Conference}, pp.~22487--22497, 2025.

\bibitem{radford2021learning}
A.~Radford, J.~W. Kim, C.~Hallacy, A.~Ramesh, G.~Goh, S.~Agarwal, G.~Sastry, A.~Askell, P.~Mishkin, J.~Clark, {\em et~al.}, ``Learning transferable visual models from natural language supervision,'' in {\em International Conference on Machine Learning}, pp.~8748--8763, PMLR, 2021.

\bibitem{alayrac2022flamingo}
J.-B. Alayrac, J.~Donahue, P.~Luc, A.~Miech, I.~Barr, Y.~Hasson, K.~Lenc, A.~Mensch, K.~Millican, M.~Reynolds, {\em et~al.}, ``Flamingo: a visual language model for few-shot learning,'' {\em Advances in neural information processing systems}, vol.~35, pp.~23716--23736, 2022.

\bibitem{rombach2021highresolution}
R.~Rombach, A.~Blattmann, D.~Lorenz, P.~Esser, and B.~Ommer, ``High-resolution image synthesis with latent diffusion models,'' 2021.

\bibitem{li2024dreamscene}
H.~Li, H.~Shi, W.~Zhang, W.~Wu, Y.~Liao, L.~Wang, L.-h. Lee, and P.~Y. Zhou, ``Dreamscene: 3d gaussian-based text-to-3d scene generation via formation pattern sampling,'' in {\em European Conference on Computer Vision}, pp.~214--230, Springer, 2024.

\bibitem{chen2023scenedreamer}
Z.~Chen, G.~Wang, and Z.~Liu, ``Scenedreamer: Unbounded 3d scene generation from 2d image collections,'' {\em IEEE transactions on pattern analysis and machine intelligence}, vol.~45, no.~12, pp.~15562--15576, 2023.

\bibitem{liu2023visual}
H.~Liu, C.~Li, Q.~Wu, and Y.~J. Lee, ``Visual instruction tuning,'' {\em Advances in neural information processing systems}, vol.~36, pp.~34892--34916, 2023.

\bibitem{shridhar2022cliport}
M.~Shridhar, L.~Manuelli, and D.~Fox, ``Cliport: What and where pathways for robotic manipulation,'' in {\em Conference on robot learning}, pp.~894--906, PMLR, 2022.

\bibitem{li2024llara}
X.~Li, C.~Mata, J.~Park, K.~Kahatapitiya, Y.~S. Jang, J.~Shang, K.~Ranasinghe, R.~Burgert, M.~Cai, Y.~J. Lee, {\em et~al.}, ``Llara: Supercharging robot learning data for vision-language policy,'' {\em arXiv preprint arXiv:2406.20095}, 2024.

\bibitem{black2024pi0}
K.~Black, N.~Brown, D.~Driess, A.~Esmail, M.~Equi, C.~Finn, N.~Fusai, L.~Groom, K.~Hausman, B.~Ichter, {\em et~al.}, ``pi0: A vision-language-action flow model for general robot control,'' {\em arXiv preprint arXiv:2410.24164}, 2024.

\bibitem{galaxeag0}
G.~Team, ``Galaxea g0: Open-world dataset and dual-system vla model,'' {\em arXiv preprint arXiv:2509.00576v1}, 2025.

\bibitem{isaacsim}
{NVIDIA}, ``{Isaac Sim}.''

\bibitem{isaaclab}
M.~Mittal {\em et~al.}, ``Orbit: A unified simulation framework for interactive robot learning environments,'' {\em IEEE Robotics and Automation Letters}, 2023.

\bibitem{xiang2020sapien}
F.~Xiang, Y.~Qin, K.~Mo, Y.~Xia, H.~Zhu, F.~Liu, M.~Liu, H.~Jiang, Y.~Yuan, H.~Wang, {\em et~al.}, ``Sapien: A simulated part-based interactive environment,'' in {\em Proceedings of the IEEE/CVF conference on computer vision and pattern recognition}, pp.~11097--11107, 2020.

\bibitem{rlbench}
S.~James, Z.~Ma, D.~Rovick~Arrojo, and A.~J. Davison, ``Rlbench: The robot learning benchmark \& learning environment,'' {\em IEEE Robotics and Automation Letters}, 2020.

\bibitem{gong2023arnold}
R.~Gong, J.~Huang, Y.~Zhao, H.~Geng, X.~Gao, Q.~Wu, W.~Ai, Z.~Zhou, D.~Terzopoulos, S.-C. Zhu, B.~Jia, and S.~Huang, ``Arnold: A benchmark for language-grounded task learning with continuous states in realistic 3d scenes,'' in {\em Proceedings of the IEEE/CVF International Conference on Computer Vision (ICCV)}, pp.~20483--20495, October 2023.

\bibitem{liu2023libero}
B.~Liu, Y.~Zhu, C.~Gao, Y.~Feng, Q.~Liu, Y.~Zhu, and P.~Stone, ``Libero: Benchmarking knowledge transfer for lifelong robot learning,'' {\em Advances in Neural Information Processing Systems}, 2023.

\bibitem{yu2020meta}
T.~Yu, D.~Quillen, Z.~He, R.~Julian, K.~Hausman, C.~Finn, and S.~Levine, ``Meta-world: A benchmark and evaluation for multi-task and meta reinforcement learning,'' in {\em Conference on robot learning}, PMLR, 2020.

\bibitem{mu2025robotwin}
Y.~Mu, T.~Chen, Z.~Chen, S.~Peng, Z.~Lan, Z.~Gao, Z.~Liang, Q.~Yu, Y.~Zou, M.~Xu, {\em et~al.}, ``Robotwin: Dual-arm robot benchmark with generative digital twins,'' in {\em Proceedings of the Computer Vision and Pattern Recognition Conference}, 2025.

\bibitem{chen2025robotwin}
T.~Chen, Z.~Chen, B.~Chen, Z.~Cai, Y.~Liu, Q.~Liang, Z.~Li, X.~Lin, Y.~Ge, Z.~Gu, {\em et~al.}, ``Robotwin 2.0: A scalable data generator and benchmark with strong domain randomization for robust bimanual robotic manipulation,'' {\em arXiv preprint arXiv:2506.18088}, 2025.

\bibitem{taomaniskill3}
S.~Tao, F.~Xiang, A.~Shukla, Y.~Qin, X.~Hinrichsen, X.~Yuan, C.~Bao, X.~Lin, Y.~Liu, and T.~kai Chan~et al., ``Maniskill3: Gpu parallelized robotics simulation and rendering for generalizable embodied ai,'' {\em Robotics: Science and Systems}, 2025.

\bibitem{li2024behavior}
C.~Li {\em et~al.}, ``Behavior-1k: A human-centered, embodied ai benchmark with 1,000 everyday activities and realistic simulation,'' {\em arXiv preprint arXiv:2403.09227}, 2024.

\bibitem{kim2024openvla}
M.~J. e.~a. Kim, ``Openvla: An open-source vision-language-action model,'' in {\em Proceedings of The 8th Conference on Robot Learning}, 2025.

\bibitem{liu2023llm+}
B.~Liu, Y.~Jiang, X.~Zhang, Q.~Liu, S.~Zhang, J.~Biswas, and P.~Stone, ``Llm+ p: Empowering large language models with optimal planning proficiency,'' {\em arXiv preprint arXiv:2304.11477}, 2023.

\bibitem{gu2024conceptgraphs}
Q.~Gu {\em et~al.}, ``Conceptgraphs: Open-vocabulary 3d scene graphs for perception and planning,'' in {\em 2024 IEEE International Conference on Robotics and Automation (ICRA)}, pp.~5021--5028, IEEE, 2024.

\bibitem{cui2024anyskill}
J.~Cui, T.~Liu, N.~Liu, Y.~Yang, Y.~Zhu, and S.~Huang, ``Anyskill: Learning open-vocabulary physical skill for interactive agents,'' in {\em Proceedings of the IEEE/CVF conference on computer vision and pattern recognition}, pp.~852--862, 2024.

\bibitem{huagensim2}
P.~Hua, M.~Liu, A.~Macaluso, Y.~Lin, W.~Zhang, H.~Xu, and L.~Wang, ``Gensim2: Scaling robot data generation with multi-modal and reasoning llms,'' in {\em 8th Annual Conference on Robot Learning}, 2024.

\bibitem{robogen}
Y.~Wang, Z.~Xian, F.~Chen, T.-H. Wang, Y.~Wang, K.~Fragkiadaki, Z.~Erickson, D.~Held, and C.~Gan, ``Robogen: Towards unleashing infinite data for automated robot learning via generative simulation,'' {\em arXiv preprint arXiv:2311.01455}, 2023.

\bibitem{wang2023gensim}
L.~Wang, Y.~Ling, Z.~Yuan, M.~Shridhar, C.~Bao, Y.~Qin, B.~Wang, H.~Xu, and X.~Wang, ``Gensim: Generating robotic simulation tasks via large language models,'' {\em arXiv preprint arXiv:2310.01361}, 2023.

\bibitem{wang2023robogen}
Y.~Wang, Z.~Xian, F.~Chen, T.-H. Wang, Y.~Wang, K.~Fragkiadaki, Z.~Erickson, D.~Held, and C.~Gan, ``Robogen: Towards unleashing infinite data for automated robot learning via generative simulation,'' 2023.

\bibitem{katara2024gen2sim}
P.~Katara, Z.~Xian, and K.~Fragkiadaki, ``Gen2sim: Scaling up robot learning in simulation with generative models,'' in {\em 2024 IEEE International Conference on Robotics and Automation (ICRA)}, IEEE, 2024.

\bibitem{ha2023scaling}
H.~Ha, P.~Florence, and S.~Song, ``Scaling up and distilling down: Language-guided robot skill acquisition,'' in {\em Conference on Robot Learning}, PMLR, 2023.

\bibitem{robocasa2024}
S.~Nasiriany, A.~Maddukuri, L.~Zhang, A.~Parikh, A.~Lo, A.~Joshi, A.~Mandlekar, and Y.~Zhu, ``Robocasa: Large-scale simulation of everyday tasks for generalist robots,'' in {\em Robotics: Science and Systems}, 2024.

\bibitem{gu2023maniskill2}
J.~Gu, F.~Xiang, X.~Li, Z.~Ling, X.~Liu, T.~Mu, Y.~Tang, S.~Tao, X.~Wei, Y.~Yao, X.~Yuan, P.~Xie, Z.~Huang, R.~Chen, and H.~Su, ``Maniskill2: A unified benchmark for generalizable manipulation skills,'' in {\em International Conference on Learning Representations}, 2023.

\bibitem{deng2025graspvla}
S.~Deng, M.~Yan, S.~Wei, H.~Ma, Y.~Yang, J.~Chen, Z.~Zhang, T.~Yang, X.~Zhang, H.~Cui, {\em et~al.}, ``Graspvla: a grasping foundation model pre-trained on billion-scale synthetic action data,'' {\em arXiv preprint arXiv:2505.03233}, 2025.

\bibitem{mees2022calvin}
O.~Mees, L.~Hermann, E.~Rosete-Beas, and W.~Burgard, ``Calvin: A benchmark for language-conditioned policy learning for long-horizon robot manipulation tasks,'' {\em IEEE Robotics and Automation Letters (RA-L)}, 2022.

\bibitem{tang2023industreal}
B.~Tang, M.~A. Lin, I.~Akinola, A.~Handa, G.~S. Sukhatme, F.~Ramos, D.~Fox, and Y.~Narang, ``Industreal: Transferring contact-rich assembly tasks from simulation to reality,'' {\em arXiv preprint arXiv:2305.17110}, 2023.

\bibitem{handa2023dextreme}
A.~Handa, A.~Allshire, V.~Makoviychuk, A.~Petrenko, R.~Singh, J.~Liu, D.~Makoviichuk, K.~Van~Wyk, A.~Zhurkevich, B.~Sundaralingam, {\em et~al.}, ``Dextreme: Transfer of agile in-hand manipulation from simulation to reality,'' in {\em 2023 IEEE International Conference on Robotics and Automation (ICRA)}, 2023.

\bibitem{akkaya2019solving}
I.~Akkaya {\em et~al.}, ``Solving rubik's cube with a robot hand,'' {\em arXiv preprint arXiv:1910.07113}, 2019.

\bibitem{yu2024natural}
A.~Yu, A.~Foote, R.~Mooney, and R.~Mart{\'\i}n-Mart{\'\i}n, ``Natural language can help bridge the sim2real gap,'' {\em arXiv preprint arXiv:2405.10020}, 2024.

\bibitem{raffel2020exploring}
C.~Raffel, N.~Shazeer, A.~Roberts, K.~Lee, S.~Narang, M.~Matena, Y.~Zhou, W.~Li, and P.~J. Liu, ``Exploring the limits of transfer learning with a unified text-to-text transformer,'' {\em Journal of machine learning research}, vol.~21, no.~140, pp.~1--67, 2020.

\bibitem{reimers-2019-sentence-bert}
N.~Reimers and I.~Gurevych, ``Sentence-bert: Sentence embeddings using siamese bert-networks,'' in {\em Proceedings of the 2019 Conference on Empirical Methods in Natural Language Processing}, Association for Computational Linguistics, 11 2019.

\bibitem{douze2024faiss}
M.~Douze, A.~Guzhva, C.~Deng, J.~Johnson, G.~Szilvasy, P.-E. Mazaré, M.~Lomeli, L.~Hosseini, and H.~Jégou, ``The faiss library,'' 2024.

\bibitem{johnson2019billion}
J.~Johnson, M.~Douze, and H.~J{\'e}gou, ``Billion-scale similarity search with {GPUs},'' {\em IEEE Transactions on Big Data}, vol.~7, no.~3, pp.~535--547, 2019.

\bibitem{ma2024eureka}
Y.~J. Ma, W.~Liang, G.~Wang, D.-A. Huang, O.~Bastani, D.~Jayaraman, Y.~Zhu, L.~Fan, and A.~Anandkumar, ``Eureka: Human-level reward design via coding large language models,'' in {\em ICRL}, 2024.

\bibitem{liang2022code}
J.~Liang, W.~Huang, F.~Xia, P.~Xu, K.~Hausman, B.~Ichter, P.~Florence, and A.~Zeng, ``Code as policies: Language model programs for embodied control,'' {\em arXiv preprint arXiv:2209.07753}, 2022.

\bibitem{openeqa}
A.~Majumdar {\em et~al.}, ``Openeqa: Embodied question answering in the era of foundation models,'' in {\em Proceedings of the IEEE/CVF conference on computer vision and pattern recognition}, 2024.

\bibitem{osada2002trimeshsampling}
R.~Osada, T.~Funkhouser, B.~Chazelle, and D.~Dobkin, ``Shape distributions,'' {\em ACM Transactions on Graphics}, 2002.

\bibitem{sundaralingam2023curobo}
B.~Sundaralingam, S.~K.~S. Hari, A.~Fishman, C.~Garrett, K.~Van~Wyk, V.~Blukis, A.~Millane, H.~Oleynikova, A.~Handa, F.~Ramos, {\em et~al.}, ``curobo: Parallelized collision-free minimum-jerk robot motion generation,'' {\em arXiv preprint arXiv:2310.17274}, 2023.

\bibitem{metaflow}
N.~O.~S. Platform, ``{Metaflow}.'' \url{https://github.com/Netflix/metaflow}, 2019.

\bibitem{guo2025deepseek}
D.~Guo, D.~Yang, H.~Zhang, J.~Song, R.~Zhang, R.~Xu, Q.~Zhu, S.~Ma, P.~Wang, X.~Bi, {\em et~al.}, ``Deepseek-r1: Incentivizing reasoning capability in llms via reinforcement learning,'' {\em arXiv preprint arXiv:2501.12948}, 2025.

\bibitem{papineni2002bleu}
K.~Papineni, S.~Roukos, T.~Ward, and W.-J. Zhu, ``Bleu: a method for automatic evaluation of machine translation,'' in {\em Proceedings of the 40th annual meeting of the Association for Computational Linguistics}, pp.~311--318, 2002.

\bibitem{DP3}
Y.~Ze, G.~Zhang, K.~Zhang, C.~Hu, M.~Wang, and H.~Xu, ``3d diffusion policy: Generalizable visuomotor policy learning via simple 3d representations,'' {\em arXiv preprint arXiv:2403.03954}, 2024.

\bibitem{act}
T.~Z. Zhao, V.~Kumar, S.~Levine, and C.~Finn, ``Learning fine-grained bimanual manipulation with low-cost hardware,'' {\em arXiv preprint arXiv:2304.13705}, 2023.

\bibitem{kingma2014adam}
D.~P. Kingma, ``Adam: A method for stochastic optimization,'' {\em arXiv preprint arXiv:1412.6980}, 2014.

\end{thebibliography}
\bibliographystyle{ieeetr}
\clearpage

%%%%%%%%%%%%%%%%%%%%%%%%%%%%%%%%%%%%%%%%%%%%%%%%%%%%%%%%%%%%%%%%%%%%%%%%%%%%%%%%

\begin{strip}
\vspace{-24mm}
    \centering
    \includegraphics[width=\textwidth]{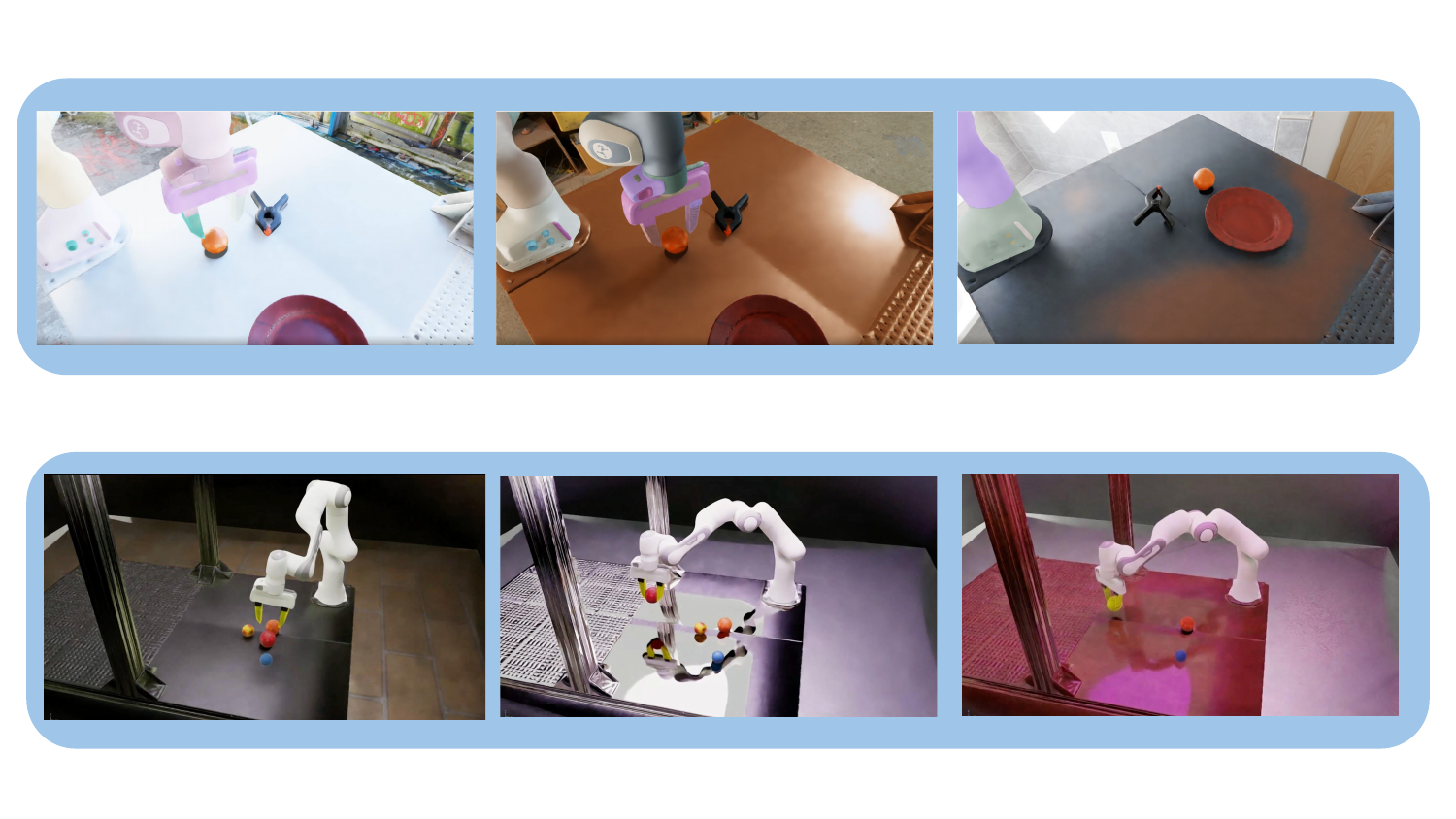}
    \vspace{-14mm}
    \captionof{figure}{Our system can automatically randomize object poses and textures}
    \label{fig:randomization}
\end{strip}
\section*{APPENDICES}
\subsection{Simulation Environments} 

\subsubsection{Rendering}
We demonstrate domain randomization capabilities of our system in \autoref{fig:randomization}.

\subsubsection{Assets}
\begin{figure*}[t]
\centering
\includegraphics[width=\linewidth]{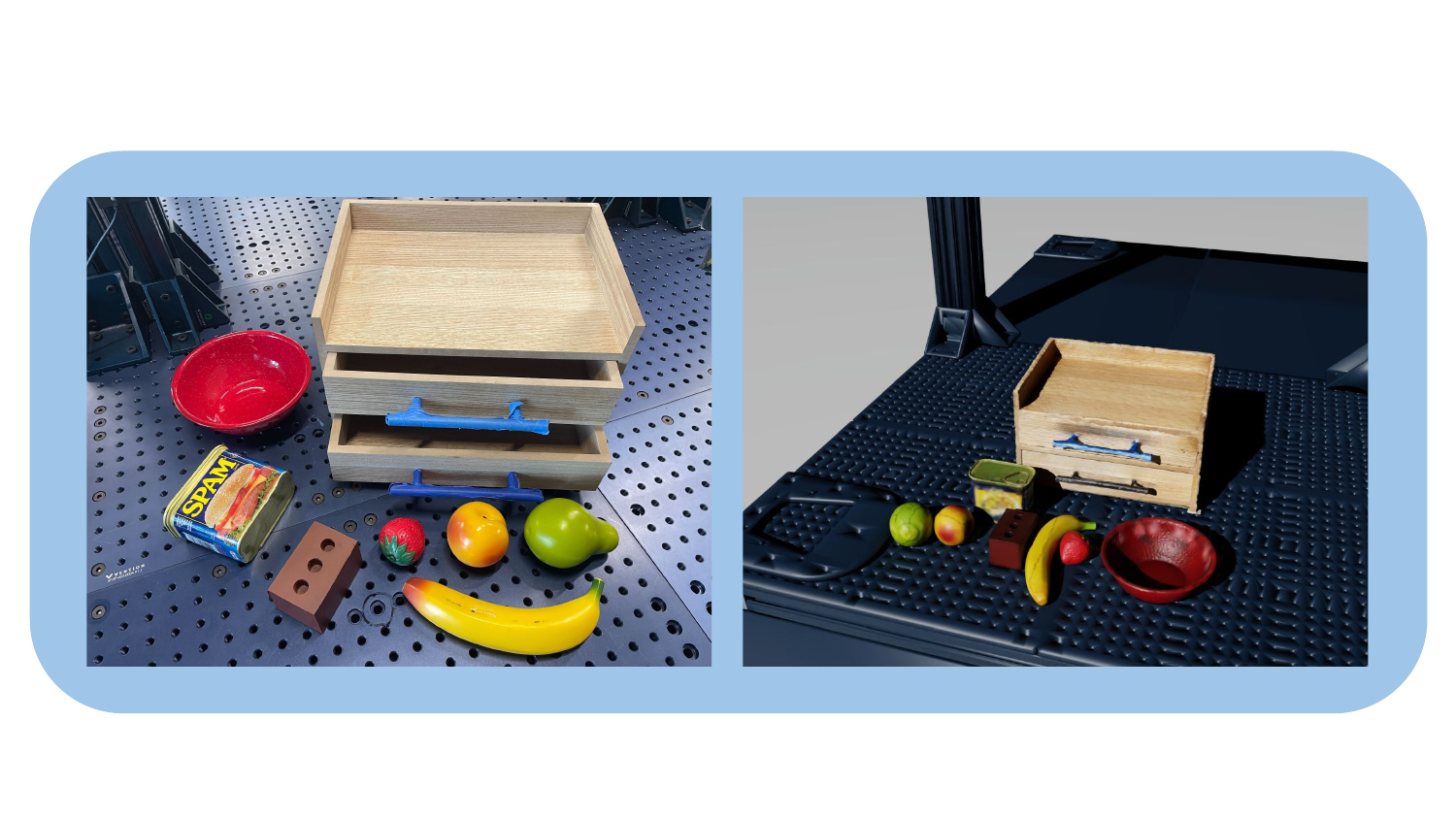}
\vspace{-15mm}
\caption{A subset of the simulated assets (right) compared to real-world assets (left).}
\label{fig:assets}
\end{figure*}

We display part of our simulated assets and real world assets in \autoref{fig:assets}. For the drawer, we manually create the URDF file and its associated links after scanning with an iPhone.

\subsection{Task Generation}\label{sec:appendix_taskgen}
\subsubsection{Object Database}\label{sec:appendix_object_database}
\begin{figure*}[!ht]
\centering
\includegraphics[width=0.22\textwidth]{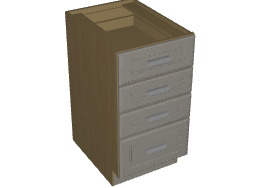}
\includegraphics[width=0.22\textwidth]{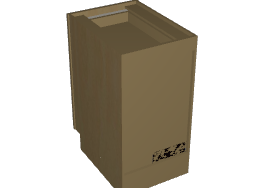}
\includegraphics[width=0.22\textwidth]{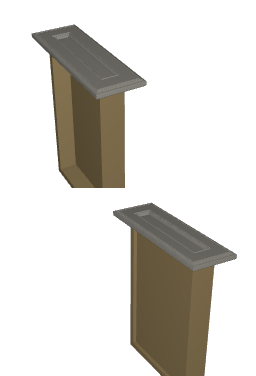}
\includegraphics[width=0.22\textwidth]{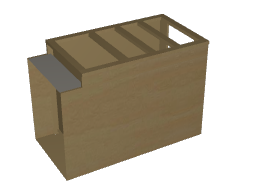}

\begin{tcolorbox}[colback=gray!5!white,colframe=gray!75!black]
\textbf{\textit{Object Metadata:}}
\begin{verbatim}
{
    'name': 'storage_furniture',
    '_name_emb': <name language embedding omitted>,
    'desc': 'The object is a storage furniture 
        piece featuring a rectangular design made from wood. 
        It includes a total of four parts: one body and three drawers, each equipped with handles. 
        The overall color is a light wood finish with neutral drawer fronts, 
        providing a clean and functional appearance.', 
    '_desc_emb': <desc language embedding omitted>, 
    'shape': 'Rectangular, vertical storage unit with multiple drawers', 
    '_shape_emb': <shape language embedding omitted>, 
    'material': 'Wood', 
    '_material_emb': <material language embedding omitted>, 
    'bbox_desc': '1.1m long, 0.8m wide, and 1.5m tall.', 
    '_bbox_desc_emb': <bbox language embedding omitted>, 
    'pattern': 'Smooth surface with no distinct patterns', 
    '_pattern_emb': <pattern language embedding omitted>, 
    '_type': 'object', 
    'bbox': (
        (-0.5500339754288982, -0.385874013080157, -0.8268710066308561), 
        (0.5771890770736514, 0.4125760342546297, 0.7034910114080105)), 
    'color': 'Light wood finish for the body; neutral color for the drawer fronts'
}
\end{verbatim}

\end{tcolorbox}
\caption{Object Database Sample: Multi-view, Multi-part Rendering and VLM-labeled metadata} \label{fig:object_database_sample}
\end{figure*}

We build the object database with VLM. \autoref{fig:object_database_sample} and \autoref{fig:object_database_sample_continued} shows examples of our multi-view, multi-part rendering and the metadata labeled by VLM.

\begin{figure*}[!ht]
\centering
\begin{tcolorbox}[colback=gray!5!white,colframe=gray!75!black]
\textbf{\textit{Object Metadata Continued (drawer part of the entire object):}}
\begin{verbatim}
{
    'label': 'drawer', 
     'desc': 'The drawer is a rectangular component of the storage furniture, made from light brown wood.
        It features a solid color with a smooth surface. This drawer is designed to slide in and out, 
        providing convenient storage space.', 
    'shape': 'rectangular', 
    'material': 'wood',
    'bbox_desc': '0.8m long, 0.3m wide, and 0.9m tall.', 
    'pattern': 'solid', '
    'name': 'link_1', 
    'joints': ['joint_1', 'joint_6'], 
    'bbox': 
        ((-0.4042590431785981, -0.16426906793320312, -0.3997560500405213), 
        (0.37755600785708515, 0.09633603205617348, 0.5206800517408057)), 
    'color': 'light brown', 
    'children': {
        'drawer_front_29': 
        'desc': 'drawer front', 
        'material': 'paint', 
        'bbox_desc': '0.0cm long, 0.4m wide, and 5.6cm tall.', 
        'color_desc': 'light beige',
        'pattern': 'solid', 
        'texture': 'This material is a light beige paint with a solid pattern.', 
        'bbox': ((-0.5028430304480387, -0.20997298721734534, -0.061841028196256004),
        (-0.5028430104844965, 0.23667601504659727, -0.006091010623107836)), 
        'color': '#fcf1e4',
        <... more children parts>
    }
\end{verbatim}

\end{tcolorbox}
\caption{Object Database Sample (Continued)} \label{fig:object_database_sample_continued}
\end{figure*}

\subsubsection{Sample Prompt for \ours{}} 
Here are some examples of \ours{} code generation prompt. Task generation prompts are shown in \autoref{fig:taskGeneration}. Success checker prompts are shown in \autoref{fig:successChecker} and \autoref{fig:successChecker2} . \TAMP{} policy prompts are shown in \autoref{fig:viprPrompt} abd \autoref{fig:ViprPrompt2}. Observation state generation prompts are in \autoref{fig:composeState}. Reward function prompts are in \autoref{fig:rewardfunctionprompt} and \autoref{fig:rewardfunctionPromptC}. 

%##################################################################################################

\begin{figure*}[t]
\centering
\begin{tcolorbox}[width=\textwidth, colback=gray!5!white, colframe=gray!75!black]
\textbf{\textit{Task Generation:}}\\
You are a robotics research scientist responsible for developing manipulation skills for a robot.
Given the following objects in the scene in a table-top environment, please propose some ycb pick and place 3obj tasks for training robot policies.
You should also follow the format requirements:\\
1. You are expected to generate a ".json" format answer to describe each task. No other words outside .json part.\\
2. You should consider the capability of the robot.\\
3. Assume the positive z-axis is the up direction, positive x-axis is the forward direction, and the negative y-axis is the right direction, from the base of the robot.\\
4. Assume the robot is placed at (0, 0, 0).\\
5. Object ids should follow the numbers in the list of objects (which starts from 1).\\
6. Label the id of relevant objects (which will be manipulated, or will be considered in environment state) in the generated configuration.\\

Example:\\
Example 0\\
List of objects:\\
1. Apple\\
2. Banana\\
3. Cup\\
Environment:\\
Table-top manipulation environment. The table is placed in front of the robot.\\
Robot:\\
Degrees of Freedom: 7 DoF\\
Payload: Up to 3 kg \\
Reach: 855 mm \\
Precision: ±0.1 mm

\begin{verbatim}
Proposed 4 tasks:
{
    "tasks":[
        {
            "task_description": "Pick up the apple and place it next to the banana.",
            "relevant_object_ids": [1, 2]
        },
        {
            "task_description": "Pick up the banana and place it in between the apple and the cup.",
            "relevant_object_ids": [1, 2, 3]
        },
        {
            "task_description": "Move the apple and place it behind the banana.",
            "relevant_object_ids": [1, 2]
        },
        {
            "task_description": "Place the apple to the right of the banana.",
            "relevant_object_ids": [1, 2]
        }
    ]
}
\end{verbatim}

\end{tcolorbox}
\caption{Task Generation Prompt}\label{fig:taskGeneration}
\end{figure*}

%##################################################################################################

%##################################################################################################
% \definecolor{codeblue}{rgb}{0.25,0.5,0.5}
% \definecolor{codekw}{rgb}{0.85, 0.18, 0.50}
% \definecolor{keywordgreen}{rgb}{0,0.6,0}
% \lstset{
%   backgroundcolor=\color{gray!10},
%   basicstyle=\fontsize{8pt}{8.5pt}\ttfamily\selectfont,
%   columns=fullflexible,
%   breaklines=true,
%   captionpos=b,
%   commentstyle=\fontsize{7.5pt}{7.5pt}\color{codeblue},
%   keywords = {AI, Human}, 
%   keywordstyle = {\textbf},
%   caption={Prompt for AnyTask success generation},
%   % label={lst:diamond}
% }
% \begin{lstlisting}[language=Python] 

\begin{figure*}[!ht]
\centering
\begin{tcolorbox}[colback=gray!5!white,colframe=gray!75!black]
\textbf{\textit{Success Checker:}}\\
You are a robotics research scientist responsible for developing manipulation skills for a robot. Given the following objects and the robot manipulation task, please write a Python method called "check\_success()" to indicate if an agent finishes the task successfully, and rephase the task description that exactly matches your check success function, especially for some numerical values specified in the function but not in the original description.

You should give your response in json format {"check\_success": <your code>, "rephrased\_task\_description": <>}\\
You should also follow the format requirements:\\
1. You are expected to generate Python code for "check\_success".\\
2. You should consider the capability of the robot.\\
3. Assume the positive z-axis is the up direction, positive x-axis is the forward direction, and the negative y-axis is the right direction, from the base of the robot.\\
4. Assume the robot is placed at (0, 0, 0).\\
5. Please keep the object id consistent with the provided list of objects.\\
6. Write the function that is compatible with vectorized environments, i.e. variables are batched arrays (tensors). This is important for bool operations.\\
You are provided with following APIs to use\\

\begin{verbatim}
{
    "type":"function",
    "function":{
        "name":"is_grasping_object",
        "description":"Check if the robot is gripping the queried object (part).    
            Args:        
            env (Any): simulation environment
            object_id (int): object id
            part_name (str): name of the object part. Defaults to "" -- no object part specified.
            If the object is an articulated object and the `part_name` is empty or not found,
            return the gathered results of its parts by `or`.
            If the object is a rigid object, ignore the `part_name`.
            env_ids (torch.Tensor | None): if specified, return the information of env_ids only.
            Returns:
            torch.Tensor: True if the robot is gripping the object (part)"}
}
\end{verbatim}
~\\
<More API descriptions omitted>\\
~\\
In addition, you are allow to use common functions in pytorch. Don't use other functions.\\
This is an example:\\
Environment:\\
Table-top manipulation environment. You can assume there will be a table. The closer right corner of the table is aligned with the robot. The table is 0.8m wide and 0.8m long.\\
Robot:\\
Degrees of Freedom: 7 DoF\\
Payload: Up to 3 kg\\
Reach: 855 mm\\
Precision: ±0.1 mm\\
List of objects:\\
1. Apple\\
2. Banana\\
3. Cup\\
Task description:\\
Pick up an apple.\\
Please generate the success check function:\\

\end{tcolorbox}
\caption{Success Checker Prompt}\label{fig:successChecker}
\end{figure*}
%##################################################################################################

\begin{figure*}[!ht]
\centering
\begin{tcolorbox}[colback=gray!5!white,colframe=gray!75!black]
\textbf{\textit{Success Checker Continued:}}\\
% [\{
"check\_success":
\begin{verbatim}
    # return a bool tensor 
    def check_success(env) -> torch.Tensor:    
        apple_id = 1    
        robot_reach = 0.8 # approximation of the robot reach
        success_grasping = is_grasping_object(env, apple_id)    
        # positions are represented by batched (x, y, z) coordinates  
        # the height is z-axis
        apple_height = get_object_position(env, apple_id)[:, 2]
        apple_initial_height = get_object_init_position(env, apple_id)[:, 2]    
        success_lifting = (apple_height - apple_initial_height) > (robot_reach * 0.5) 
        # lift apple up of half of the robot reach    
        success_lifting = success_lifting & (apple_height > 0) # apple is above the robot base    
        return success_grasping & success_lifting
\end{verbatim}
",\\
"rephrased\_task\_description": "pick up an apple and lift it up of half of the robot reach"\\
% \}\\
====\\
Now here is the information for you:\\
Environment:\\
Table-top manipulation environment. You can assume there will be a table. The closer left corner of the table surface is aligned with the robot base (0, 0, 0). Assume the positive z-axis is the up direction, positive x-axis is the forward direction, and the negative y-axis is the right direction, from the base of the robot. Your work space is [0.15, 0.65] in x,  [-0.68, -0.2] in y, in meters. This is for reference and you don\'t need to add a table.\\
Robot:\\
Degrees of Freedom: 7 DoF Payload: Up to 3 kg Reach: 855 mm Precision: ±0.1 mm\\
List of objects:\\
1. \{'object\_name': 'skillet lid', 'object\_description': 'skillet lid'\}\\
2. \{'object\_name': 'marbles', 'object\_description': 'marbles'\}\\
3. \{'object\_name': 'cups', 'object\_description': 'cups'\}\\
Task description:\\
Pick up the marbles and place them next to the skillet lid.\\
Please generate the success check function:

\textbf{\textit{Success Checker Response}}:\\
\{    "check\_success": "

...\\
...\\
\}
\end{tcolorbox}
\caption{Success Checker Prompt Continued}\label{fig:successChecker2}
\end{figure*}

\definecolor{codeblue}{rgb}{0.25,0.5,0.5}
\definecolor{codekw}{rgb}{0.85, 0.18, 0.50}
\definecolor{keywordgreen}{rgb}{0,0.6,0}
\lstset{
  backgroundcolor=\color{gray!10},
  basicstyle=\fontsize{8pt}{8.5pt}\ttfamily\selectfont,
  columns=fullflexible,
  breaklines=true,
  captionpos=b,
  commentstyle=\fontsize{7.5pt}{7.5pt}\color{codeblue},
  keywords = {AI, Human}, 
  keywordstyle = {\textbf},
  % caption={Prompt for AnyTask scripted policy generation},
  % label={lst:diamond}
}

\begin{figure*}[!ht]
\centering
\begin{tcolorbox}[colback=gray!5!white,colframe=gray!75!black]
\textbf{\textit{ \TAMP{} Policy Prompt:}}

You are a robotics research scientist responsible for developing manipulation skills for a robot. Given the following objects, the robot manipulation task, and the corresponding success checker code, please write a Python method called "scripted\_policy()" that executes a scripted policy to finish the task. The scripted policy is composed by a sequence of pre-defined skills with associated parameters. In order to better store the steps you solve the problem by scripted policy, use `log\_step\_description` API to note down your explanation of the sub step in natural language sentences. Refer to this API\'s docstring for the usage guide. You are encouraged to give the exact values when necessary and the information is easy to get from environment APIs. Python\'s f-string is useful. Based on the scripted policy you generated, please also label the ordering of the objects being manipulated. Your final output format of this request is **json format** like \\
\{"scripted\_policy": ..., "object\_manipulation\_order": [\{"object\_id": <object id, should be consistent with the information in given list of objects>, "part\_name": <if the object is articulated object, specify the part it wants to manipulate. Use null if there is no>, "joint\_name": <if the object is articulated object, specify the joint that related to manipulating that part, if the joint is not a fixed joint. Use null if there is not.>\}]\}, \\
where `scripted\_policy` contains your code and `object\_manipulation\_order` contains the object manipulation order.\\
You should also follow the format requirements:\\
1. You are expected to generate a Python format answer. No other words outside Python part.\\
2. You should consider the capability of the robot.\\
3. Assume the positive z-axis is the up direction, positive x-axis is the forward direction, and the negative y-axis is the right direction, from the base of the robot.\\
4. Assume the robot is placed at (0, 0, 0).\\
5. Please keep the object id consistent with the provided list of objects.\\
6. Write the function that is compatible with vectorized environments, i.e. variables are batched arrays (tensors). This is important for bool operations.\\
7. Try to match your scripted policy with the success checker so that your policy can success. \\
8. No pre-grasp steps.\\
9. For tasks involving operating articulated object, it's good to use the API to check the current joint position and joint limits, then decide the parameters for skills, but you don't need to use a loop. You can determine the value and execute it by one shot. Avoid guessing how much you should operate. Also, don't use the axis provided in joint attribute to determine the operation direction of the joint because the orientation of the object changes. Use API instead.\\
You are provided with following skill APIs:
\begin{verbatim}
{
    "type":"function",
    "function":{
        "name":"move_to",
        "description":"Move the robot end effector to target position and orientation, 
        with the gripper status unchanged.
        Args:
            env (Any): simulation environment
            target_position (torch.Tensor): target position to move, shape (N, 3)
            target_orientation (torch.Tesnor | None): target orientation in quat, shape (N, 4), or None
            Defaults to None (keep the current orientation).
            gripper_open (bool): `True` to open the gripper, `False` to close the gripper. 
            Default to True.",
    "parameters":{"type":"object","properties":{
        "env":{"type":"Any"},
        "target_position":{"type":"torch.Tensor"},
        "target_orientation":{"type":"torch.Tensor | None"},
        "gripper_open":{"type":"bool"}},
        "required":["env","target_position","target_orientation"]}}
\end{verbatim}
<More skill APIs omitted>
\end{tcolorbox}
\caption{\TAMP{} Policy Prompt}\label{fig:viprPrompt}
\end{figure*}

\begin{figure*}[!ht]
\centering
\begin{tcolorbox}[colback=gray!5!white,colframe=gray!75!black]
\textbf{\textit{\TAMP{}  Policy Prompt Continued:}}\\
In addition, you are allow to use common functions in pytorch.\\
This is an example:\\
Environment: Table-top manipulation environment. You can assume there will be a table. The closer right corner of the table is aligned with the robot. The table is 0.8m wide and 0.8m long.\\
Robot: Degrees of Freedom: 7 DoF Payload: Up to 3 kg Reach: 855 mm Precision: ±0.1 mm\\
List of objects:\\
1. Apple 2. Banana 3. Cup\\
Task description:\\
Place the apple next to the banana.\\
<Success checker code omitted>\\
Please generate the scripted policy:\\
\begin{verbatim}
{"scripted_policy": "
    import torch
    ...
    ...
    object_manipulation_order": {"object_id": 1, "part_name": null, "joint_name": null}
}
\end{verbatim}
====\\
Now here is the information for you:
Environment:\\
Table-top manipulation environment. You can assume there will be a table.
The closer left corner of the table surface is aligned with the robot base (0, 0, 0). 
Assume the positive z-axis is the up direction, positive x-axis is the forward direction, 
and the negative y-axis is the right direction, from the base of the robot. Your work space is [0.15, 0.65] in x,  [-0.68, -0.2] in y, in meters. This is for reference and you don\'t need to add a table.\\
Robot:\\
Degrees of Freedom: 7 DoF\\
Payload: Up to 3 kg\\
Reach: 855 mm\\
Precision: ±0.1 mm\\
List of objects:\\
1. {'object\_name': 'skillet lid', 'object\_description': 'skillet lid'}\\
2. {'object\_name': 'marbles', 'object\_description': 'marbles'}\\
3. {'object\_name': 'cups', 'object\_description': 'cups'}\\
Task description:\\
Pick up the skillet lid and place it on top of the cups. The task is considered successfully completed if the robot is no longer grasping the skillet lid, the lid is positioned within 0.05 meters horizontally (in the x-y plane) of the cups, and its height is at least 0.02 meters above that of the cups.\\
Please generate the scripted policy:\\
\textbf{\textit{\TAMP{} Policy Response:}} \\
\begin{verbatim}
import torch
...
...
"object_manipulation_order": [{"object_id": 1, "part_name": null, "joint_name": null}]
\end{verbatim}
\end{tcolorbox}
\caption{\TAMP{} prompt continued }\label{fig:ViprPrompt2}
\end{figure*}

\definecolor{codeblue}{rgb}{0.25,0.5,0.5}
\definecolor{codekw}{rgb}{0.85, 0.18, 0.50}
\definecolor{keywordgreen}{rgb}{0,0.6,0}
\lstset{
  backgroundcolor=\color{gray!10},
  basicstyle=\fontsize{8pt}{8.5pt}\ttfamily\selectfont,
  columns=fullflexible,
  breaklines=true,
  captionpos=b,
  commentstyle=\fontsize{7.5pt}{7.5pt}\color{codeblue},
  keywords = {AI, Human}, 
  keywordstyle = {\textbf},
  % caption={Prompt for AnyTask state generation},
  % label={lst:diamond}
}

\begin{figure*}[!ht]
\centering
\begin{tcolorbox}[colback=gray!5!white,colframe=gray!75!black]
\textbf{\textit{Compose State Prompt:}}

You are a robotics research scientist responsible for developing manipulation skills for a robot. Given the following objects, the robot manipulation task, and the corresponding success checker code, please write a Python function compose\_state() to compose state information required to successfully train an RL policy. You should also follow the format requirements:\\
1. You are expected to generate a Python format answer. No other words outside Python part.\\
2. You should consider the capability of the robot.\\
3. Assume the positive z-axis is the up direction, positive x-axis is the forward direction, and the negative y-axis is the right direction, from the base of the robot.\\
4. Assume the robot is placed at (0, 0, 0).\\
5. Object ids should follow the numbers in the list of objects (which starts from 1).\\
6. We already provide the state information of robot itself. You will be responsible for object states or robot-object states.\\
7. At least cover all relevant objects.\\
8. You can compose any privileged state information that is helpful for RL, as long as only the provided APIs and common operations are used.\\
9. Write the function that is compatible with vectorized environments, i.e. variables are batched arrays (tensors). This is important for bool operations.\\
10. The definition of the compose\_state() is `def compose\_state(env) -> dict[str, torch.Tensor]:`, where `env` is the simulation environment and the return value is a dictionary where the key is the name of the state and the value is the state information (in torch.Tensor). Don\'t use duplicated keys.\\
11. You are not able to access check\_success() directly.\\
You are provided with following APIs to use\\

<APIs are omitted>\\

=====\\
In addition, you are allow to use common functions in pytorch.\\
Example:\\
List of objects:\\
1. Apple 2. Banana 3. Cup\\
Environment:\\
Table-top manipulation environment. The table is placed in front of the robot.\\
Robot: Degrees of Freedom: 7 DoF Payload: Up to 3 kg Reach: 855 mm Precision: ±0.1 mm\\
Task description:\\
Pick up the apple and place it next to the banana.\\
Relevant object ids: [1, 2]\\
Please generate the `compose\_state` function:
\begin{verbatim}
def compose_state(env)-> dict[str, torch.Tensor]:
    # relevant objects
    apple_id = 1
    banana_id = 2
    # end-effector to apple distance
    eef_to_apple = object_to_robot_eef_distance(env, apple_id)
    # end-effector to banana distance
    eef_to_banana = object_to_robot_eef_distance(env, banana_id)
    # apple to banana
    apple_to_banana = distance(get_object_position(env, apple_id), 
        get_object_position(env, banana_id))
    return {
        "eef_to_apple": eef_to_apple, 
        "eef_to_banana": eef_to_banana, 
        "apple_to_banana": apple_to_banana
    }
\end{verbatim}
=====\\

\end{tcolorbox}
\caption{Compose state prompt}\label{fig:composeState}
\end{figure*}

\begin{figure*}[!ht]
\centering
\begin{tcolorbox}[colback=gray!5!white,colframe=gray!75!black]
\textbf{\textit{Compose State Continued:}} \\
Now here is the task:\\
List of objects:\\
1. \{'object\_name': 'skillet lid', 'object\_description': 'skillet lid'\}\\
2. \{'object\_name': 'marbles', 'object\_description': 'marbles'\}\\
3. \{'object\_name': 'cups', 'object\_description': 'cups'\}\\
Environment:\\
Table-top manipulation environment. You can assume there will be a table. The closer left corner of the table surface is aligned with the robot base (0, 0, 0). Assume the positive z-axis is the up direction, positive x-axis is the forward direction, and the negative y-axis is the right direction, from the base of the robot. Your work space is [0.15, 0.65] in x,  [-0.68, -0.2] in y, in meters. This is for reference and you don't need to add a table.\\
Robot:\\
Degrees of Freedom: 7 DoF\\
Payload: Up to 3 kg\\
Reach: 855 mm\\
Precision: ±0.1 mm\\
Task description:\\
Pick up the marbles and place them next to the skillet lid such that the marbles, once released by the robot, are within 0.15 meters of the skillet lid in Euclidean distance and have a vertical (z-axis) difference smaller than 0.1 meters.\\
Relevant object ids: [1, 2]\\
Success checker code:\\
~\\
\textbf{\textit{Compose State Response:}}\\
\begin{verbatim}
import torch
def check_success(env) -> torch.Tensor:
    # Object ids
    skillet_lid_id = 1
    marbles_id = 2
    # Thresholds
    placement_threshold = 0.15
    # max horizontal distance for marbles to be considered next to skillet lid (in meters)
    vertical_threshold = 0.1    # max allowed difference in z-axis (in meters)
    # Get current positions of the objects; shape: (N, 3)
    marbles_pos = get_object_position(env, marbles_id)
    skillet_lid_pos = get_object_position(env, skillet_lid_id)
    # Compute Euclidean distance between marbles and skillet lid
    dist = distance(marbles_pos, skillet_lid_pos)
    # Check if horizontal placement is within threshold
    placement_success = dist < placement_threshold
    # Check vertical alignment: the z difference should be small (within vertical_threshold)
    z_diff = torch.abs(marbles_pos[:, 2] - skillet_lid_pos[:, 2])
    vertical_alignment = z_diff < vertical_threshold
    # Ensure that the marbles are not currently being grasped
    not_grasping = ~is_grasping_object(env, marbles_id, "")
    # Overall success: marbles placed next to the skillet lid 
    # (both horizontal and vertical conditions) and released
    success = placement_success & vertical_alignment & not_grasping
    return success
\end{verbatim}
\end{tcolorbox}
\caption{Compose state prompt continued}\label{fig:composeState}
\end{figure*}
%##################################################################################################

%##################################################################################################
\definecolor{codeblue}{rgb}{0.25,0.5,0.5}
\definecolor{codekw}{rgb}{0.85, 0.18, 0.50}
\definecolor{keywordgreen}{rgb}{0,0.6,0}
\lstset{
  backgroundcolor=\color{gray!10},
  basicstyle=\fontsize{8pt}{8.5pt}\ttfamily\selectfont,
  columns=fullflexible,
  breaklines=true,
  captionpos=b,
  commentstyle=\fontsize{7.5pt}{7.5pt}\color{codeblue},
  keywords = {AI, Human}, 
  keywordstyle = {\textbf},
  % caption={Prompt for AnyTask reward generation},
  % label={lst:diamond}
}
\begin{figure*}[!ht]
\centering
\begin{tcolorbox}[colback=gray!5!white,colframe=gray!75!black]
\textbf{\textit{Reward Function:}}

You are a robotics research scientist responsible for developing manipulation skills for a robot. Given the following objects, the robot manipulation task, and the corresponding success checker code, please write a Python method called "reward\_function()" to assign rewards to an RL agent that is learning this manipulation task. You should also follow the format requirements:\\
1. You are expected to generate a Python format answer. No other words outside Python part.\\
2. You should consider the capability of the robot.\\
3. Assume the positive z-axis is the up direction, positive x-axis is the forward direction, and the negative y-axis is the right direction, from the base of the robot.\\
4. Assume the robot is placed at (0, 0, 0).\\
5. Please keep the object id consistent with the provided list of objects.\\
6. Write the function that is compatible with vectorized environments, i.e. variables are batched arrays (tensors). This is important for bool operations.\\
7. Try your best to design the reward function with the reference of success condition (success checker). But you are not able to access check\_success() directly.\\
You are provided with following APIs to use:\\
~\\
<Environment APIs are omitted>\\
~\\

In addition, you are allow to use common functions in pytorch.
This is an example:\\
Environment:\\
Table-top manipulation environment. You can assume there will be a table. The closer right corner of the table is aligned with the robot. The table is 0.8m wide and 0.8m long.\\
Robot:\\
Degrees of Freedom: 7 DoF\\
Payload: Up to 3 kg\\
Reach: 855 mm\\
Precision: ±0.1 mm\\
List of objects:\\
1. Apple\\
2. Banana\\
3. Cup\\
Task description:\\
Pick up an apple.\\
Please generate the reward function:\\
\begin{verbatim}
def reward_function(env):
    objects = get_object_list(env)
    # apple's object id is 1
    apple_id = 1
    # query the distance to robot end effector
    reward = zero_rewards(env)
    env_is_grasping_object = is_grasping_object(env, apple_id)
    # success grasping
    reward[env_is_grasping_object] += 100
    # apple is being picked up
    current_apple_position = get_object_position(env, apple_id)
    init_apple_position = get_object_initial_position(env, apple_id)
    total_delta_z = current_apple_position[:, 2] - init_apple_position[:, 2]
    reward[env_is_grasping_object & (total_delta_z > 0.3)] += 10000
    return reward
\end{verbatim}
====\\
\end{tcolorbox}
\caption{Reward function prompt.} \label{fig:rewardfunctionprompt}
\end{figure*}

\begin{figure*}[!ht]
\centering
\begin{tcolorbox}[colback=gray!5!white,colframe=gray!75!black]
\textbf{\textit{Reward Function Continued:}} \\
Now here is the information for you:\\
Environment:\\
Table-top manipulation environment. You can assume there will be a table. The closer left corner of the table surface is aligned with the robot base (0, 0, 0). Assume the positive z-axis is the up direction, positive x-axis is the forward direction, and the negative y-axis is the right direction, from the base of the robot. Your work space is [0.15, 0.65] in x,  [-0.68, -0.2] in y, in meters. This is for reference and you don't need to add a table.\\
Robot:\\
Degrees of Freedom: 7 DoF\\
Payload: Up to 3 kg\\
Reach: 855 mm\\
Precision: ±0.1 mm\\
List of objects:\\
1. \{'object\_name': 'skillet lid', 'object\_description': 'skillet lid'\}\\
2. \{'object\_name': 'marbles', 'object\_description': 'marbles'\}\\
3. \{'object\_name': 'cups', 'object\_description': 'cups'\}\\
Task description:\\
Pick up the skillet lid and place it on top of the cups. The task is considered successfully completed if the robot is no longer grasping the skillet lid, the lid is positioned within 0.05 meters horizontally (in the x-y plane) of the cups, and its height is at least 0.02 meters above that of the cups.\\
<Success checker code omitted>\\
Please generate the reward function:\\
~\\
\textbf{\textit{Reward function response:}}
\begin{verbatim}
def reward_function(env):
    import torch
    ...
    ...
    return reward
\end{verbatim}

\end{tcolorbox}
\caption{Reward function prompt continued.} \label{fig:rewardfunctionPromptC}
\end{figure*}
%##################################################################################################

% --- First table: lifting tasks ---
\begin{table*}[thb]
    \centering
    \caption{Task Table - Example Lifting Tasks}
    \label{tab:task_table_lifting}
    \resizebox{\textwidth}{!}{%
    \begin{tabular}{c c p{0.7\textwidth}}
        \toprule
        Task ID & Task Type & Description \\
        \midrule
        3 & ycb\_lifting\_1obj & Lift the tennis ball from the table surface by grasping it and holding it aloft without placing or stacking it. \\
        7 & ycb\_lifting\_1obj & Grasp the banana from the table and smoothly lift it upward to a height of approximately 10 cm above its initial position, holding it steady for a few seconds. This task focuses solely on the lifting motion without placing or stacking actions. \\
        9 & ycb\_lifting\_1obj & Grasp the apple and lift it vertically upward by approximately 0.15 m above its initial position. \\
        10 & ycb\_lifting\_1obj & Pick up and lift the phillips screwdriver from the table. \\
        16 & ycb\_lifting\_1obj & Grasp the medium clamp from the table and lift it vertically upward until it is completely off the table surface, then hold it steady. \\
        25 & ycb\_lifting\_1obj & Grasp the flat screwdriver from the table and lift it upward without placing it back down. \\
        27 & ycb\_lifting\_1obj & Pick up the foam brick from its current position on the table and lift it vertically upward, holding it in the air without placing or stacking it on any other surface. \\
        29 & ycb\_lifting\_1obj & Lift the flat screwdriver from the table and hold it in the air at a fixed height. \\
        14 & ycb\_lifting\_2obj & Grasp the tennis ball from the table and lift it vertically upward, ensuring the ball is elevated away from the table surface. \\
        15 & ycb\_lifting\_2obj & Grasp the strawberry from the table and lift it gently upward, maintaining a secure grip during the lift. \\
        0 & ycb\_lifting\_3obj & Lift the large clamp by raising it vertically off the table by approximately 15 cm. \\
        9 & ycb\_lifting\_3obj & Grasp the orange and lift it vertically upward from its starting position on the table. \\
        17 & ycb\_lifting\_3obj & Lift the phillips screwdriver by grasping it from the table and lifting it straight up to verify the grasp. \\
        \bottomrule
    \end{tabular}%
    }
    \label{tab:tasktablelifting}
\end{table*}

% --- Second table: pick and place tasks ---
\begin{table*}[thb]
    \centering
    \caption{Task Table - Example Pick and Place Tasks}
    \label{tab:task_table_pick_and_place}
    \resizebox{\textwidth}{!}{%
    \begin{tabular}{c c p{0.7\textwidth}}
        \toprule
        Task ID & Task Type & Description \\
        \midrule
        0 & ycb\_pick\_and\_place\_1obj & Pick up the phillips screwdriver from its current location on the table and place it at a target location (e.g., at x = 0.40 m, y = -0.45 m) within the workspace. \\
        14 & ycb\_pick\_and\_place\_1obj & Pick up the apple and place it at a target location on the table (e.g., at x = 0.5, y = -0.4), ensuring that the apple is well within the robot's reachable workspace. \\
        16 & ycb\_pick\_and\_place\_1obj & Pick up the apple and place it at a central location within the workspace (e.g., x: 0.4, y: -0.45) to ensure clearance from the table edges. \\
        23 & ycb\_pick\_and\_place\_1obj & Pick up the potted meat can and place it at the front center of the workspace on the table. \\
        2 & ycb\_pick\_and\_place\_2obj & Pick up the fork and place it to the left of the plum on the table. \\
        18 & ycb\_pick\_and\_place\_2obj & Pick up the marbles and place them directly in front of the padlock on the table. \\
        25 & ycb\_pick\_and\_place\_2obj & Pick up the golf ball and place it in front of the padlock on the table. \\
        37 & ycb\_pick\_and\_place\_2obj & Pick up the power drill and place it to the right of the bleach cleanser on the table. \\
        0 & ycb\_pick\_and\_place\_3obj & Pick up the tennis ball and place it next to the pear. \\
        \bottomrule
    \end{tabular}%
    }
    \label{tab:tasktablepicking}
\end{table*}

% --- Third table: pushing tasks ---
\begin{table*}[thb]
    \centering
    \caption{Task Table - Example Pushing Tasks}
    \label{tab:task_table_pushing}
    \resizebox{\textwidth}{!}{%
    \begin{tabular}{c c p{0.7\textwidth}}
        \toprule
        Task ID & Task Type & Description \\
        \midrule
        5 & ycb\_pushing\_1obj & Push the tuna fish can using a steady forward push to a designated target location near the center of the workspace, ensuring controlled contact during motion. \\
        10 & ycb\_pushing\_1obj & Push the nine hole peg test object from its initial position to the center of the workspace (approximately x=0.4, y=-0.45) to simulate a controlled translational motion across the table. \\
        17 & ycb\_pushing\_1obj & Push the banana from its initial position to a target location near the center of the workspace (e.g., around x=0.4, y=-0.45) using a gentle linear pushing motion. \\
        19 & ycb\_pushing\_1obj & Push the mustard bottle across the table surface from its current position near the left corner towards the center of the workspace, ensuring controlled contact and avoiding excessive force that might tip the bottle. \\
        20 & ycb\_pushing\_1obj & Push the flat screwdriver along the table: starting from its initial position within the workspace, push it in a straight line along the positive x direction (forward) to a target location near the forward edge of the workspace. This task requires precise control to ensure the tool remains aligned while being pushed. \\
        21 & ycb\_pushing\_1obj & Push the pear gently along the positive x-axis (forward direction) until it reaches the center of the workspace. \\
        24 & ycb\_pushing\_1obj & Push the skillet lid towards the center of the workspace at approximately (0.4, -0.44), ensuring the movement is smooth and along the table surface. \\
        39 & ycb\_pushing\_2obj & Push the spoon such that it is repositioned to the left of the sponge (i.e., with a higher y value, closer to y = -0.2) to create a clear spatial separation between the two objects. \\
        18 & ycb\_pushing\_3obj & Push the pear along the x-axis so that it reaches a target zone near (x=0.50, y=-0.45) on the table. \\
        22 & ycb\_pushing\_3obj & Push the bowl from its current position to a target area, ensuring the marbles remain close by. \\
        \bottomrule
    \end{tabular}%
    }
    \label{tab:tasktablepushing}
\end{table*}

\subsubsection{Environment APIs}\label{sec:appendix_envapi}
Environment APIs are in \autoref{tab:appendix_env_apis}. Detailed argument definitions are available in our code.
\begin{table*}[tbp]
    \centering
    \caption{List Environment APIs}
    \begin{tabular}{ll}
    \toprule
    %TODO: sort them
      API & Description \\
      \midrule
      \texttt{get\_object\_position()}   &  Object position \\
      \texttt{get\_object\_initial\_position()}   &  Object position at the beginning of the episode \\
      \texttt{get\_object\_last\_position()}   & Object position at last simulation step \\
      \texttt{get\_object\_rotation()}   & Object orientation \\
      \texttt{get\_object\_initial\_rotation()}   &  Object orientation at the beginning of the episode \\
      \texttt{get\_object\_last\_rotation()}   & Object orientation at last simulation step \\
      \texttt{get\_contact\_grasp\_position\_external()}   & Grasp pose wrt an object \\
      \texttt{check\_no\_collision()}   & Check if the objects are colliding \\
      \texttt{get\_workspace\_limit()}   & Obtain the workspace limits \\
      \texttt{sample\_position\_in\_workspace()}   & Sample a position within the workspace (for placing the object) \\
      \texttt{check\_within\_workspace()}   & Check if the object is within the workspace \\
      \texttt{add\_object\_to\_env()}   & Place an object into the scene \\
      \texttt{get\_object\_list()}   & Get the list of objects \\
      \texttt{point\_to\_object\_distance()}   & Distance between the point and an object \\
      \texttt{get\_robot\_joint\_positions()}   &  Robot joint positions \\
      \texttt{get\_robot\_eef\_orientation()}   &  Robot eef orientation \\
      \texttt{get\_robot\_eef\_position()}   &  Robot eef position \\
      \texttt{get\_robot\_eef\_last\_position()}   & Robot eef position at last simulation step\\
      \texttt{get\_robot\_eef\_initial\_position()}   &  Robot eef position at the beginning of the episode\\
      \texttt{distance()}   &  Distance between two objects \\
      \texttt{object\_to\_robot\_eef\_distance()}   & Distance between the object and the robot eef \\
      \texttt{point\_to\_robot\_eef\_distance()}   & Distance between a point and the robot eef \\
      \texttt{is\_grasping\_object()}   & Is the robot grasping the object \\
      \texttt{success\_grasp()}   & Has the robot successfully grasped the object  \\
      \texttt{maintain\_grasp()}   & Is the robot maintaining the grasping \\
      \texttt{drop\_grasp()}   &  Has the robot dropped the object \\
      \texttt{get\_object\_bbox()}   & Object bounding box \\
      \texttt{get\_articulated\_obj\_joint\_state()}   & State of a joint in an articulated object \\
      \texttt{get\_articulated\_obj\_joint\_limits()}   & Limits of a joint in an articulated object \\
      \texttt{get\_articulated\_obj\_joint\_operating\_direction()}   & Operating direction of a joint in an articulated object \\
      \texttt{get\_articulated\_obj\_part\_pos()}   & Position of a part in an articulated object \\
      \texttt{get\_articulated\_obj\_part\_rotation()}   & Orientation of a part in an articulated object \\
      \texttt{get\_articulated\_obj\_part\_bbox()}   &  Bounding box of a part in an articulated object \\
      \texttt{sample\_orientation\_facing\_towards\_robot()}   & Sample an orientation that makes the object face towards the robot \\
      \texttt{zero\_rewards()}   & Create a zero reward tensor \\
      \texttt{log\_step\_description()}   & Log the language description of a scripted policy step \\
      \texttt{make\_a\_tensor()}   & Make a tensor \\
    \bottomrule
    \end{tabular}\label{tab:appendix_env_apis}
\end{table*}

\subsection{Trajectory Generation} \label{sec:appendix_trajectory_generation}

\subsubsection{Skill APIs}
Please find skill APIs in \autoref{tab:appendix_skill_apis}. Detailed argument definitions are available in our code.
\begin{table*}[tbp]
    \centering
    \caption{List of Skill APIs}
    \begin{tabular}{ll}
    \toprule
    API & Description \\
    \midrule
      \texttt{reset\_robot()}   & Move the robot to home position \\
      \texttt{pick()}   & Pick up an object \\
      \texttt{close\_gripper()}   & Close gripper \\
      \texttt{open\_gripper()}   &  Open gripper \\
      \texttt{place()}   & Place object \\
      \texttt{push\_or\_pull\_object\_to\_xy()}   & Push or pull the object \\
      \texttt{move\_to()}   & Move the robot to a position \\
    \bottomrule
    \end{tabular}
    \label{tab:appendix_skill_apis}
\end{table*}

\subsubsection{\ours{} Agents examples}
In this section, we present several examples generated by our system using \ours{} agents. A \TAMP{} example is illustrated in \autoref{fig:vipr_policy}, while a \RL{} example is provided in \autoref{fig:rl_policy}.

\lstset{
    language=Python,
    basicstyle=\ttfamily\scriptsize, % Changed to \scriptsize (even smaller than \footnotesize)
    breaklines=true,
    breakatwhitespace=true, % Helps prevent breaking words in half
    showstringspaces=false,
    commentstyle=\color{gray},
    keywordstyle=\color{blue}\bfseries,
    stringstyle=\color{red},
    lineskip=0pt,        % Tighten line spacing
    belowskip=0pt,        % Remove space below the listing
    aboveskip=0pt,        % Remove space above the listing
    columns=flexible,    % <<< CHANGE: Use flexible columns for better line breaking
    keepspaces=true
}
\begin{figure*}[!ht]
\centering
% *** REMOVED \resizebox COMMAND ***
\begin{tcolorbox}[
    title=\TAMP{},
    colback=gray!5!white,
    colframe=gray!75!black,
    sharp corners,
    boxrule=0.5pt,
    enlarge top by=1mm,
    enlarge bottom by=1mm,
    width=\textwidth % Ensure tcolorbox itself is full width
]
\begin{lstlisting}
import torch

def scripted_policy(env):
    # Object IDs
    softball_id = 1
    clamp_id = 2
    # Get initial positions for the softball and the robot's end effector
    softball_position = get_object_position(env, softball_id)  # shape (N,3)
    robot_eef_position = get_robot_eef_position(env)
    # Log the state before picking up the softball
    log_step_description(
        env,
        before_step={
            "template": "Softball is at {softball_position} and robot EEF is at {robot_eef_position} before grasping.",
            "kwargs": {
                "softball_position": softball_position,
                "robot_eef_position": robot_eef_position
            }
        },
        action_todo={
            "template": "Approach the softball to grasp it.",
            "kwargs": {}
        }
    )
    # Approach Phase: move above the softball
    hover_offset = torch.tensor([[0.0, 0.0, 0.1]], device=softball_position.device)  # 10 cm above
    above_softball = softball_position + hover_offset
    move_to(env, target_position=above_softball, target_orientation=None, gripper_open=True)
    # Move down to the softball to grasp it
    move_to(env, target_position=softball_position, target_orientation=None, gripper_open=True)
    # Grasp the softball by closing the gripper
    close_gripper(env)
    # Lift Phase: lift the softball by moving back to the above position
    move_to(env, target_position=above_softball, target_orientation=None, gripper_open=False)
    # Determine target placement: place softball on top of extra large clamp
    clamp_position = get_object_position(env, clamp_id)  # shape (N,3)
    # Set target such that the horizontal (x, y) centers align with clamp, and elevate by additional 0.05 m
    target_position = torch.cat([clamp_position[:, :2], clamp_position[:, 2:3] + 0.05], dim=1)
    # Log the state before placing the softball
    log_step_description(
        env,
        before_step={
            "template": "Preparing to place softball: clamp is at {clamp_position}, target placement is {target_position}.",
            "kwargs": {
                "clamp_position": clamp_position,
                "target_position": target_position
            }
        },
        action_todo={
            "template": "Move to target placement and release the softball.",
            "kwargs": {}
        }
    )
    # Move above the target placement
    above_target = target_position + hover_offset
    move_to(env, target_position=above_target, target_orientation=None, gripper_open=False)
    # Lower the softball to the target placement on top of the clamp
    move_to(env, target_position=target_position, target_orientation=None, gripper_open=False)
    # Release Phase: open the gripper to drop the softball
    open_gripper(env)
    # Retract: move back up to above the target placement
    move_to(env, target_position=above_target, target_orientation=None, gripper_open=True)
    # Log the final state after placing the softball
    log_step_description(
        env,
        after_step={
            "template": "After placing, softball is at {softball_pos} and clamp is at {clamp_pos}.",
            "kwargs": {
                "softball_pos": get_object_position(env, softball_id),
                "clamp_pos": clamp_position
            }
        }
    )
\end{lstlisting}
\end{tcolorbox}

\caption{Generated policy demonstrating a sequence of robotic actions to pick up a softball and place it on a clamp.}\label{fig:vipr_policy}
\end{figure*}

\lstset{
    language=Python,
    % *** CHANGE HERE: Font size reduced from 8pt to 7pt ***
    basicstyle=\fontsize{6pt}{6.5pt}\ttfamily\selectfont, 
    columns=fullflexible,
    breaklines=true,
    breakatwhitespace=true,
    backgroundcolor=\color{gray!10},
    % Removed internal listing caption settings
    commentstyle=\fontsize{6.5pt}{6.5pt}\color{codeblue}, % Comment style adjusted to match
    keywords = {AI, Human}, 
    keywordstyle = {\textbf},
    lineskip=-0.8pt,    % Aggressively tighten lines even further
    aboveskip=0pt,
    belowskip=0pt,
}

\begin{figure*}[!ht]
\centering
\begin{tcolorbox}[
    title=\RL{},
    colback=gray!5!white,
    colframe=gray!75!black,
    sharp corners,
    boxrule=0.5pt,
    enlarge top by=1mm,
    enlarge bottom by=1mm,
    width=\textwidth
]
\begin{lstlisting}[language=Python]
import torch

def scripted_policy_rl(env):
    # Object IDs
    softball_id = 1
    clamp_id = 2

    # ------------------- Part 1: Pick up the softball -------------------
    # Use contact sampling to get the grasp pose for the softball via RL-based skill
    grasp_position, grasp_orientation = get_contact_grasp_position_external(env, softball_id, part_name="")

    # Log before picking
    log_step_description(
        env,
        before_step={
            "template": "Softball grasp pose computed: position {grasp_position}, orientation {grasp_orientation}.",
            "kwargs": {"grasp_position": grasp_position, "grasp_orientation": grasp_orientation}
        },
        action_todo={
            "template": "Moving to grasp pose for softball.",
            "kwargs": {}
        }
    )

    # Move the robot's end effector to the computed grasp pose
    move_to(
        env=env,
        target_position=grasp_position,
        target_orientation=grasp_orientation,
        gripper_open=True
    )

    # Execute the RL-based picking skill for the softball
    pick_success = pick_rl(env, external_id=softball_id)

    # Log after pick
    log_step_description(
        env,
        after_step={
            "template": "RL pick executed for softball (ID {softball_id}). Success: {pick_success}.",
            "kwargs": {"softball_id": softball_id, "pick_success": pick_success}
        }
    )

    # ------------------- Part 2: Place the softball on the extra large clamp -------------------
    # Get the current position of the clamp
    clamp_position = get_object_position(env, clamp_id)

    # Define placement: center the softball on the clamp with a vertical offset to satisfy stacking constraints
    # Horizontal alignment: within 0.05m is required. We choose the clamp center.
    # Vertical offset: at least 0.03m above the clamp. We use 0.05m as a safe offset.
    placement_vertical_offset = 0.05
    target_position = clamp_position + torch.tensor([[0.0, 0.0, placement_vertical_offset]], device=clamp_position.device)

    # Define an above-placement position, by adding an extra vertical offset (e.g., 0.1 m)
    above_offset = torch.tensor([[0.0, 0.0, 0.1]], device=target_position.device)
    above_target_position = target_position + above_offset

    # Log before placing
    log_step_description(
        env,
        before_step={
            "template": "Target placement for softball on clamp: position {target_position}.",
            "kwargs": {"target_position": target_position}
        },
        action_todo={
            "template": "Moving to placement pose above the clamp.",
            "kwargs": {}
        }
    )

    # Move above the placement position (with the object in hand)
    move_to(
        env=env,
        target_position=above_target_position,
        target_orientation=None,
        gripper_open=False
    )

    # Move down to the target placement position
    move_to(
        env=env,
        target_position=target_position,
        target_orientation=None,
        gripper_open=False
    )

    # Open the gripper to release the softball
    open_gripper(env)

    # Lift the robot's end effector upward after releasing
    move_to(
        env=env,
        target_position=above_target_position,
        target_orientation=None,
        gripper_open=True
    )

    # Log after placing
    log_step_description(
        env,
        after_step={
            "template": "Softball placed on clamp. Softball position: {softball_pos}, Clamp position: {clamp_pos}.",
            "kwargs": {"softball_pos": get_object_position(env, softball_id), "clamp_pos": clamp_position}
        }
    )
\end{lstlisting}
\end{tcolorbox}
\caption{Generated \RL{} policy for the robot softball placement task. pick\_rl is the invoked RL skill. }\label{fig:rl_policy}
\end{figure*}

\subsubsection{State, Reward, Success, and Domain Randomization}
These four Python functions—$\texttt{compose\_state}$, $\texttt{reward\_function}$, $\texttt{check\_success}$, and $\texttt{domain\_randomize}$—define the environment's state representation, the task objective, the success condition, and the environment initialization for reinforcement learning. These are important components for \Eureka{}. A detailed example is in \autoref{fig:combined_rl_logic}.

\lstset{
    language=Python,
    % Font set to smallest practical size for maximum vertical fit
    basicstyle=\fontsize{5.5pt}{5.5pt}\ttfamily\selectfont, 
    columns=fullflexible,
    breaklines=true,
    breakatwhitespace=true,
    backgroundcolor=\color{gray!10},
    % Removed internal listing caption settings
    commentstyle=\fontsize{6.5pt}{6.5pt}\color{codeblue},
    keywords = {AI, Human}, 
    keywordstyle = {\textbf},
    lineskip=-0.8pt,    % Aggressively tightened line spacing
    aboveskip=0pt,
    belowskip=0pt,
}

\begin{figure*}[!ht]
\centering
\begin{tcolorbox}[
    title=\Eureka{},
    colback=gray!5!white,
    colframe=gray!75!black,
    sharp corners,
    boxrule=0.5pt,
    enlarge top by=1mm,
    enlarge bottom by=1mm,
    width=\textwidth
]
\begin{lstlisting}
def compose_state(env) -> dict[str, torch.Tensor]:
    import torch
    # Relevant object IDs for the task
    softball_id = 1
    clamp_id = 2

    # Get the current positions of the softball and the clamp
    pos_softball = get_object_position(env, softball_id)  # shape (N, 3)
    pos_clamp = get_object_position(env, clamp_id)          # shape (N, 3)
    
    # Compute horizontal (x-y plane) difference between softball and clamp
    horizontal_diff = torch.norm(pos_softball[:, :2] - pos_clamp[:, :2], dim=1, keepdim=True)
    # Compute vertical difference (z dimension) between softball and clamp
    vertical_diff = (pos_softball[:, 2:3] - pos_clamp[:, 2:3])
    
    # Compute distance from robot end-effector to the softball and to the clamp
    eef_to_softball = object_to_robot_eef_distance(env, softball_id).unsqueeze(1)  # shape (N,1)
    eef_to_clamp = object_to_robot_eef_distance(env, clamp_id).unsqueeze(1)        # shape (N,1)
    
    # Compose state dictionary with object positions and key distances
    state = {
        "softball_position": pos_softball, 
        "clamp_position": pos_clamp,
        "horizontal_diff": horizontal_diff,
        "vertical_diff": vertical_diff,
        "eef_to_softball": eef_to_softball,
        "eef_to_clamp": eef_to_clamp,
    }
    
    return state
################################################################################
def reward_function(env):
    import torch
    # object ids based on the provided list:
    # softball: 1, extra large clamp: 2
    softball_id = 1
    clamp_id = 2

    # initialize reward tensor as zero
    reward = zero_rewards(env)

    # get the current positions of the softball and the clamp (shape: (N, 3))
    pos_softball = get_object_position(env, softball_id)
    pos_clamp = get_object_position(env, clamp_id)

    # compute the horizontal (x-y) distance between the centers of softball and clamp
    horizontal_distance = torch.norm(pos_softball[:, :2] - pos_clamp[:, :2], dim=1)

    # compute the vertical difference (z-axis): softball - clamp
    vertical_diff = pos_softball[:, 2] - pos_clamp[:, 2]

    # Shaping rewards:
    # Encourage the agent to minimize the difference to the target horizontal offset (0.05 m).
    r_horizontal = 1.0 - torch.tanh(10.0 * torch.abs(horizontal_distance - 0.05))
    # Encourage lifting the softball above the clamp.
    r_vertical = torch.tanh(5.0 * (vertical_diff - 0.03))

    shaping_reward = r_horizontal + r_vertical
    reward = reward + shaping_reward

    # Bonus reward: if the softball is well stacked on the clamp per the success conditions.
    success = (horizontal_distance < 0.05) & (vertical_diff > 0.03)
    reward[success] += 10000.0

    # Additional bonus if the robot is currently grasping the softball (encourage pickup)
    grasping = is_grasping_object(env, softball_id)
    reward[grasping] += 100.0

    return reward
################################################################################
def check_success(env) -> torch.Tensor:
    import torch
    # object ids for the task
    softball_id = 1
    clamp_id = 2

    # Get current positions of the softball and the extra large clamp
    pos_softball = get_object_position(env, softball_id)  # shape (N, 3)
    pos_clamp = get_object_position(env, clamp_id)          # shape (N, 3)

    # Compute the horizontal distance (in the x-y plane) between the softball and the clamp
    horizontal_diff = torch.norm(pos_softball[:, :2] - pos_clamp[:, :2], dim=1)

    # Compute the vertical difference (z-axis). The softball should be higher than the clamp
    vertical_diff = pos_softball[:, 2] - pos_clamp[:, 2]

    # Define thresholds: horizontal difference must be less than 0.05 m and vertical difference more than 0.03 m
    success_horizontal = horizontal_diff < 0.05
    success_vertical = vertical_diff > 0.03

    # The task is successful if both conditions are met
    return success_horizontal & success_vertical
################################################################################
def domain_randomize(env, env_ids: torch.Tensor) -> torch.Tensor:
    import torch
    import numpy as np
    
    # Get workspace limits: workspace is defined for x and y.
    workspace_limit = get_workspace_limit(env)
    # Number of environments
    N = len(env_ids)
    # Initialize success flag for each environment as True
    success = torch.ones(N, device=env.device, dtype=torch.bool)
    
    # We'll assume the table-top height is fixed at 0.0 (i.e. objects are placed on the table surface).
    table_z = 0.0

    # Object ids:
    # 1: softball (target), 2: extra large clamp (support), 3: spoon (irrelevant)
    
    # Place the extra large clamp first.
    clamp_id = 2
    clamp_pos = sample_position_in_workspace(env, workspace_limit, env_ids=env_ids)
    clamp_pos[:, 2] = table_z
    success = success & add_object_to_env(env, clamp_id, clamp_pos, env_ids=env_ids)
    
    # Place the softball.
    softball_id = 1
    softball_pos = sample_position_in_workspace(env, workspace_limit, env_ids=env_ids)
    softball_pos[:, 2] = table_z
    success = success & add_object_to_env(env, softball_id, softball_pos, env_ids=env_ids)
    
    # Place the spoon (irrelevant object) at a random position.
    spoon_id = 3
    spoon_pos = sample_position_in_workspace(env, workspace_limit, env_ids=env_ids)
    spoon_pos[:, 2] = table_z
    success = success & add_object_to_env(env, spoon_id, spoon_pos, env_ids=env_ids)
    
    return success
\end{lstlisting}
\end{tcolorbox}
\caption{Basic components from \Eureka{} , including the environment $\texttt{compose\_state}$, the $\texttt{reward\_function}$, the $\texttt{check\_success}$ function, and the $\texttt{domain\_randomize}$ functions.}
\label{fig:combined_rl_logic}
\end{figure*}

\subsubsection{Object Manipulation Order Configuration}
This configuration defines the order in which objects must be manipulated to successfully complete the task. An example is shown in \autoref{fig:manipulation_order}.

\lstset{
    language=python, 
    basicstyle=\fontsize{8pt}{8.5pt}\ttfamily\selectfont, % Using 8pt font
    columns=fullflexible,
    breaklines=true,
    breakatwhitespace=true,
    backgroundcolor=\color{gray!10},
    commentstyle=\fontsize{7.5pt}{7.5pt}\color{codeblue},
    keywords = {object_id, part_name, joint_name}, % JSON keys as keywords
    keywordstyle = {\textbf\color{blue}},
    stringstyle=\color{red}, % JSON string values in red
    lineskip=0pt,    
    aboveskip=0pt,
    belowskip=0pt,
}

\begin{figure*}[!ht]
\centering
\begin{tcolorbox}[
    title=Manipulation Order,
    colback=gray!5!white,
    colframe=gray!75!black,
    sharp corners,
    boxrule=0.5pt,
    enlarge top by=1mm,
    enlarge bottom by=1mm,
    width=0.45\textwidth % Reduced width as this is a very short snippet
]
\begin{lstlisting}
"object_manipulation_order": [
        {
          "object_id": 1,
          "part_name": null,
          "joint_name": null
        },
        {
          "object_id": 2,
          "part_name": null,
          "joint_name": null
        }
      ],
\end{lstlisting}
\end{tcolorbox}
\caption{The $\texttt{object\_manipulation\_order}$ configuration, specifying that object 1 (softball) must be handled before object 2 (clamp) in the manipulation , which will be used as an input to the contact sampling.}
\label{fig:manipulation_order}
\end{figure*}

\subsection{Policy Learning} \label{sec:policy_learning_sup}
We train single task policies on our generated data and evaluate them in simulation.  For simulation evaluation, we train a single-task 3D diffusion policy~\cite{DP3} on 500 demonstrations.  The policy is conditioned on both a point cloud observation and on the robot's current end-effector position and gripper state.  We train each policy for 75,000 steps on one H100 GPU with a batch size of 1024, which takes approximately 8 hours.  We use a learning rate of 0.00005 with a cosine schedule and the Adam optimizer~\cite{kingma2014adam}.  We evaluate only the final checkpoint for each model.  Our main results required training approximately 100 policies, requiring roughly 800 GPU hours.

\subsection{Sim2Real}\label{sec:sim-to-real-appendix}
\begin{figure*}[ht]
\centering
\includegraphics[width=\linewidth]{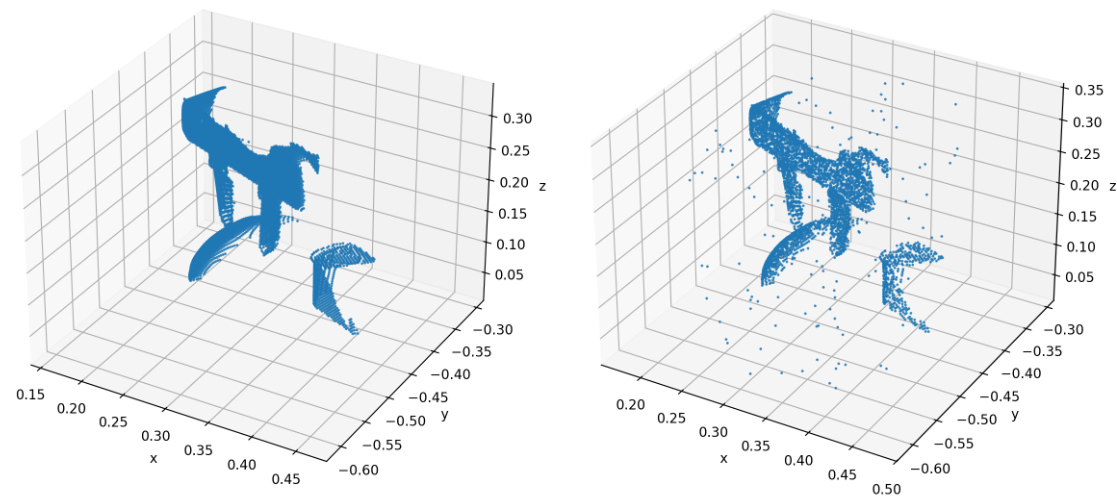}
\caption{Point-cloud augmentation for robust sim-to-real transfer.}
\label{fig:sim2real_pcd}
\end{figure*}
\begin{figure*}[t]
\centering
\includegraphics[width=\linewidth]{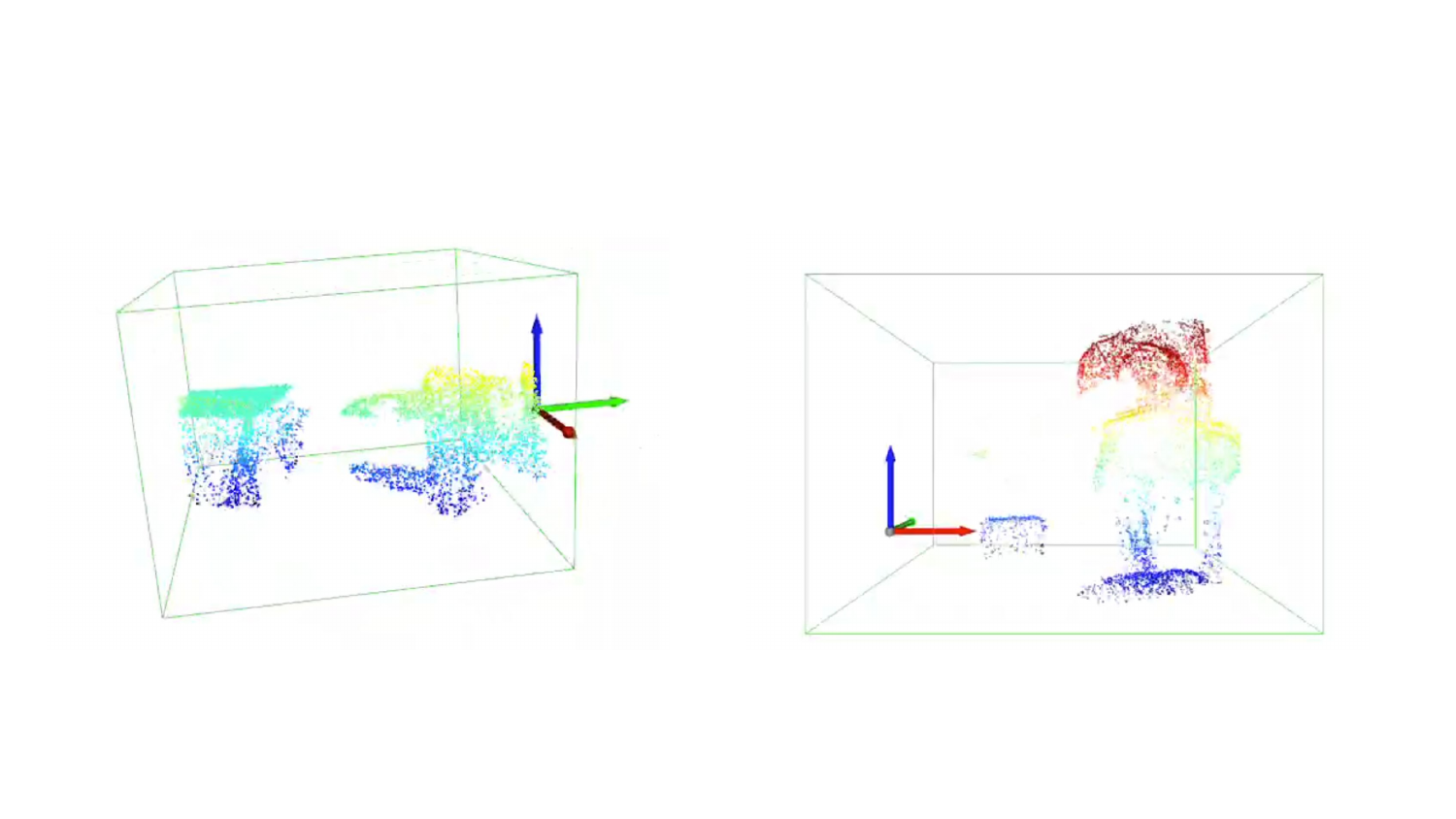}
\caption{Point-cloud observations during real-world deployment.}
\label{fig:real_pcd}
\end{figure*}
\autoref{fig:sim2real_pcd} illustrates the point-cloud augmentation strategy used for sim-to-real transfer in the banana-on-can stacking task. The left panel shows the original point cloud rendered directly from simulation, which is clean and densely structured. The right panel shows the augmented point cloud that is actually fed into policy training. This augmented input includes global position and rotation jitter, simulated flying points to mimic sensor noise and outliers, and uniform downsampling to match real-world perception sparsity. 
As a reference, \autoref{fig:real_pcd} shows the real-world observations for drawer opening and stacking tasks.
We observed that by better approximating real sensor artifacts, these augmentations improves robustness and generalization during real-world deployment.

\end{document}